\documentclass[journal]{IEEEtran}
\ifCLASSINFOpdf
\else
\fi

\usepackage{lineno,hyperref,multirow,graphicx,subfigure,enumerate,color,amsmath, epstopdf}
\usepackage{algorithm, algorithmic}

\UseRawInputEncoding

\begin{document}
%
\title{
	Research on Efficient Fuzzy Clustering Method Based on Local Fuzzy Granular balls
	}
%
%
%

\author{Jiang~Xie,
        Qiao~Deng,
        Shuyin~Xia*  ,
        Yangzhou~Zhao,
        Guoyin~Wang,
        Xinbo~Gao
}

\maketitle 

\begin{abstract}
In recent years, the problem of fuzzy clustering has been widely concerned. The membership iteration of existing methods is mostly considered globally, which has considerable problems in noisy environments, and iterative calculations for clusters with a large number of different sample sizes are not accurate and efficient. In this paper, starting from the strategy of large-scale priority, the data is fuzzy iterated using granular-balls, and the membership degree of data only considers the two granular-balls where it is located, thus improving the efficiency of iteration. The formed fuzzy granular-balls set can use more processing methods in the face of different data scenarios, which enhances the practicability of fuzzy clustering calculations.

\end{abstract}


%
\IEEEpeerreviewmaketitle

\section{Introduction}

In recent years, with the rapid development of the Internet, the era of big data has come. Data analysis plays an indispensable role in understanding various phenomena [1]. As a result, much progress has been made in the study of machine learning. Clustering is an unsupervised classification of patterns (observations, data items, or eigenvectors) by groups (clusters).[2] Data clustering is an important issue in various fields, including data mining, pattern recognition and bioinformatics. [3] There are many algorithms to solve this problem, such as the classical fuzzy C-means clustering, which is often used to segment medical images [4].

The process of dividing a collection of physical or abstract objects into multiple classes of similar objects is called clustering. A cluster generated by a cluster is a collection of data objects that are similar to objects in the same cluster and different from objects in other clusters. There are a lot of classification problems in the natural sciences and social sciences. Cluster analysis, also known as group analysis, is a statistical analysis method for studying classification problems (samples or indicators). Based on the type of application and the type of data considered, many clustering methods have been developed in various fields and are widely used in many practical applications. [5] Clustering may be the primary task for exploratory data analysis and data processing applications. Cluster analysis is the main task of grouping datasets into different object groups. To some extent, objects in the same group, or objects called clusters, are more similar to each other than objects [6] in other groups.

Cluster analysis originates from taxonomy, but clustering is not equal to classification. Clustering differs from classification in that the classes required for clustering are unknown. Cluster analysis is very rich in content, including system clustering, ordered sample clustering, dynamic clustering, fuzzy clustering, graph theory clustering, cluster prediction, and so on. Clustering is also an important concept in data mining.
Clustering is used as a preprocessing step to divide data into manageable parts as a tool for knowledge extraction. Clustering or clustering is a form of exploratory data analysis in which data is grouped into groups or subsets so that objects in each group share some similarity. The traditional calculation methods of cluster analysis are as follows:1. Partition method; 2. Hierarchical approach; 3. Density-based methods; 4. Grid-based methods; 5. Model-based approach.

Clustering can be divided into two categories based on uncertainty, hard and soft. Hard clustering strictly divides each object to be identified into a certain class, which has the nature of either one or the other. Fuzzy clustering establishes an uncertain description of the class by samples and reflects the objective world more objectively, thus becoming the mainstream of cluster analysis. Fuzzy clustering algorithm is a kind of clustering algorithm based on function optimum method. Calculus calculation technique is used to find the optimal cost function. Probability density function will be used in probability-based clustering method. To get rid of this problem, assuming an appropriate model, the vectors of fuzzy clustering algorithm can belong to multiple clusters at the same time. The method of K-means and its derivative is a hotspot in cluster research in recent years. Hard C-means clustering (HCM) is a hard cluster, while fuzzy C-means clustering (FCM) is a soft cluster in the field of k-means. [7]

FCM [8] has been shown to have better performance than HCM, which has become the most well-known and powerful method in cluster analysis. Unlike hard clustering, the FCM algorithm divides the partition of a cluster into the degree to which the data points belong to a class. [9] However, these FCM algorithms have considerable problems in noisy environments and are inaccurate for cluster iteration calculations with a large number of different sample sizes. A good clustering algorithm should be robust and can tolerate these situations that often occur in application systems. [10]
In order to solve the above problems in FCM, we propose a method based on the combination of  granular-balls and FCM to determine the initial iteration center, which not only improves the iteration efficiency, but also improves the accuracy and clustering results. So the main contributions of granular fuzzy clustering are as follows:

1. The use of granular-balls for local membership iteration can not only ensure the fuzzy attributes of data, but also greatly improve the efficiency of membership iteration.

2. The resulting fuzzy granular set can use multiple methods to complete the final clustering for different data set scenarios, such as using connect to adaptively obtain the clustering results of complex shapes; K-means can get clustering results more quickly.

3. The iterative process of the proposed method is fully adaptive, and does not require any parameter input.

\section{Related work}

\subsection{Fuzzy Sets}
Fuzzy sets appeared in a 1965 paper by Lotfi Zadeh. In 1969, Ruspini published a seminal paper that has become the basis for most fuzzy clustering algorithms. His ideas established the underlying structure of fuzzy partitions, and described and exemplified the first algorithms to achieve fuzzy partitions. The general case of fuzzy c-means models was developed by Bezdek in 1973. [11]. Senapati et al. proposed Fermat fuzzy sets, compared Fermat fuzzy sets with Pythagorean fuzzy sets and intuitionistic fuzzy sets, and found out the basic operation set of Fermat fuzzy sets. The score function and exact function of sorting Fermat fuzzy sets are also defined, and the Euclidean distance between two Fermat fuzzy sets is studied. It is also proposed in the literature to use the Fermat fuzzy TOPSIS method to solve the multi-criteria decision-making problem [12]. One of the most useful extensions to fuzzy sets for applying information uncertainty is the Fermat fuzzy set [13]. Sets such as interval-valued fuzzy sets, intuitionistic fuzzy sets, type-2 fuzzy sets, type-N fuzzy sets, hesitant fuzzy sets (HFS), dual fuzzy sets, and neutrophil sets are broad extensions of fuzzy sets that are useful in solving decision-making The problem aspect has many advantages over classical ensembles [14].

\subsection{Fuzzy Clustering}
Along with the formation, development and deepening of fuzzy set theory, RusPini first proposed the concept of fuzzy division. Taking this as the starting point and foundation, the theory and method of fuzzy clustering develop rapidly and vigorously. It generally refers to constructing a fuzzy matrix according to the attributes of the object itself, and determining the clustering relationship according to a certain degree of membership. Through fuzzy mathematics, quantitatively determine the fuzzy relationship between samples, so as to cluster objectively and accurately [15]. Fuzzy clustering analysis can be roughly divided into three categories according to the different clustering processes: fuzzy clustering based on fuzzy relationship [16], fuzzy clustering algorithm based on objective function, and fuzzy clustering algorithm based on neural network. Fuzzy cluster analysis is a classic method to provide soft partitioning of data [17]. Fuzzy cluster analysis is a mathematical method to classify things according to certain requirements when it involves fuzzy boundaries between things. In the context of fuzzy clustering, all objects share their membership to all clusters, which is inversely proportional to the distance between the object and the corresponding cluster representation [18]. Fuzzy clustering methods discover fuzzy partitions where observations can be softly assigned to multiple clusters. [19] Cluster analysis is a multivariate analysis method in mathematical statistics. It uses mathematical methods to quantitatively determine the relationship between samples, so as to objectively divide the types. The boundaries between things, some are exact, others are fuzzy. The boundary between the degree of facial resemblance in the sample population is blurred, and the boundary between cloudy and sunny weather is also blurred. When clustering involves fuzzy boundaries between things, fuzzy cluster analysis methods are used.
\subsection{Fuzzy C-means (FCM) Method}
Fuzzy C-means (FCM) is the first method to quantify the uncertainty of cluster membership by introducing fuzzy set theory into cluster groups [20]. The fuzzy C-means (FCM) algorithm is the most classic method in fuzzy clustering because of its strong fuzzy properties. FCM is a clustering method based on the fuzzy distance of the center point. In FCM, a data point is assigned to more than one cluster with a certain probability. [21]. The FCM algorithm outputs a membership degree matrix according to the minimized objective function, and the class to which the maximum membership of each data point belongs is the class to which the data point is classified. In FCM, the membership degree of a data point is iteratively updated, the number of clusters is predetermined, and all clusters have a data point with a corresponding membership degree. [22] To maximize the difference between clusters, fractional entropy can be added as a regularization function to the objective function of the FCM algorithm [23]. The fuzzy C-means algorithm is an improvement of the ordinary C-means algorithm. The ordinary C-means algorithm is rigid for data division, while FCM is a flexible fuzzy division. The simplicity of the FCM algorithm makes it very robust in most cases, and it has strong vitality in clustering algorithms. [24] Because FCM is supported by fuzzy theory, it is mainly described by fuzzy partition matrix, which breaks the limitation that each data point in hard clustering can only be divided into one category. The final output fuzzy matrix reflects all the information of the data set, so the distribution and overall characteristics of the data set can be accurately examined [25]. Although, the traditional FCM algorithm has some weaknesses, such as initializing the cluster center, determining the optimal number of clusters and being sensitive to noise, and the realization of FCM depends on the value of the fuzzy index, and the results obtained by using different indexes may be different[26 ]. The most studied is the fuzzy clustering algorithm based on the objective function, which describes the clustering problem as a constrained optimization problem, and determines the fuzzy division and clustering results of the data set by seeking the solution of the optimization problem.

Although, the traditional FCM algorithm has some weaknesses, such as initializing the cluster centers, determining the optimal number of clusters, and is susceptible to noise and outliers, because the algorithm assigns weights to all points in the same way, and cannot effectively learn from the actual data. points to distinguish noise and outliers [27] but it is still the most widely used clustering algorithm. Many validity functions suitable for FCM algorithms have been proposed in previous work, however, these traditional fuzzy set validity indices have some obvious shortcomings, such as when the number of clusters becomes large and close to the number of data points, the Appearing a monotonically decreasing trend, the fuzzy weighted index may lead to numerical instability, insufficient number of clusters, etc. [28].

The membership function is a function of the degree to which an object x belongs to set A, which is usually recorded as $ \mu_{\mathrm{A}(\mathrm{x})} $. The range of independent variables is all objects that may belong to set A, and the range of values is [0,1], that is, $ 0 \le \mu_{\mathrm{A}(\mathrm{x})} \le 1 $. $ \mu_{\mathrm{A}(\mathrm{x})} $ = 1 means that x belongs to set A completely, which is equivalent to $ \mathrm{x} \in \mathrm{A} $ on the traditional set concept. K-Means clustering is known as C-Means clustering. Its core idea is that the algorithm divides n vectors $ X_{j} $ (i = 1,2,..., n) into c classifications $ V_{i} $ (i = 1,2,..., c) and finds the cluster centers of each group so that the value function (or objective function) of the non-similarity (or distance) index is minimized. The objective function can be defined as:

\begin{equation}
\label{equ:1}
J_{m}(U, V)=\sum_{i=1}^{c} \sum_{j=1}^{n}\left\|X_{j}-V_{i}\right\|_{A}^{2}
\end{equation}

The main steps of algorithm implementation:
Step 1: Randomly determine K initial points as centroids.
Step 2: Find the nearest cluster for each data point in the dataset.
Step 3: For each cluster, compute the mean of all points in the cluster and use the mean as the centroid.
Step 4: Repeat step 2 until the cluster assignment results for any point remain unchanged.

The objective function of the FCM algorithm is

\begin{equation}
\label{equ:2}
J_{m}(U, V)=\sum_{i=1}^{c} \sum_{j=1}^{n} u_{i j}^{m}\left\|X_{j}-V_{i}\right\|_{A}^{2}
\end{equation}

Where U = [$ u_{ij} $] is the membership matrix, $ u_{ij} $ is the jth sample's membership to class i, and M is a fuzzy constant.
Constraints of the FCM algorithm: For any sample, the sum of its membership to each cluster is 1.

\begin{equation}
\label{equ:3}
\sum_{i=1}^{c} u_{i j}=1
\end{equation}

Solve U, V: To find the extreme value of the objective function under constrained conditions, we construct a new function using Lagrange multiplier method.

\begin{equation}
\label{equ:4}
F=\sum_{i=1}^{c} u_{i j}^{m}\left\|X_{j}-V_{i}\right\|^{2}+\lambda\left(\sum_{i=1}^{c} u_{i j}-1\right)
\end{equation}
among $ \lambda  $ Become a Lagrange multiplier, and find the best value for the F function under the following conditions:

\begin{equation}
\label{equ:5}
\frac{\partial F}{\partial \lambda}=\left(\sum_{i=1}^{c} u_{i j}-1\right)=0
\end{equation}

\begin{equation}
\label{equ:6}
\frac{\partial F}{\partial u_{i j}}=\left[m\left(u_{i j}\right)^{m-1}\left\|X_{j}-V_{i}\right\|^{2}-\lambda\right]=0
\end{equation}

\begin{equation}
\label{equ:7}
\frac{\partial F}{\partial v_{i}}=\sum_{j=1}^{n}\left(u_{i j}\right)^{m} x_{j}-v_{i} \sum_{i=1}^{c} u_{i j}^{m}=0
\end{equation}
Solved by the extreme condition:
\begin{equation}
\label{equ:8}
u_{i j}=\left[\sum_{k=1}^{c}\left(\frac{\left\|X_{j}-V_{i}\right\|^{2}}{\left\|X_{j}-V_{k}\right\|^{2}}\right)^{\frac{2}{m-1}}\right]
\end{equation}
\begin{equation}
\label{equ:9}
v_{i}=\frac{\sum_{j=1}^{n} u_{i j}^{m} X_{j}}{\sum_{j=1}^{n}\left(u_{i j}^{m}\right)}
\end{equation}

\begin{figure*}
	\centering

	\subfigure[]{
		\label{fig:subfig:Dataset4}
		\includegraphics[width=0.23\textwidth]{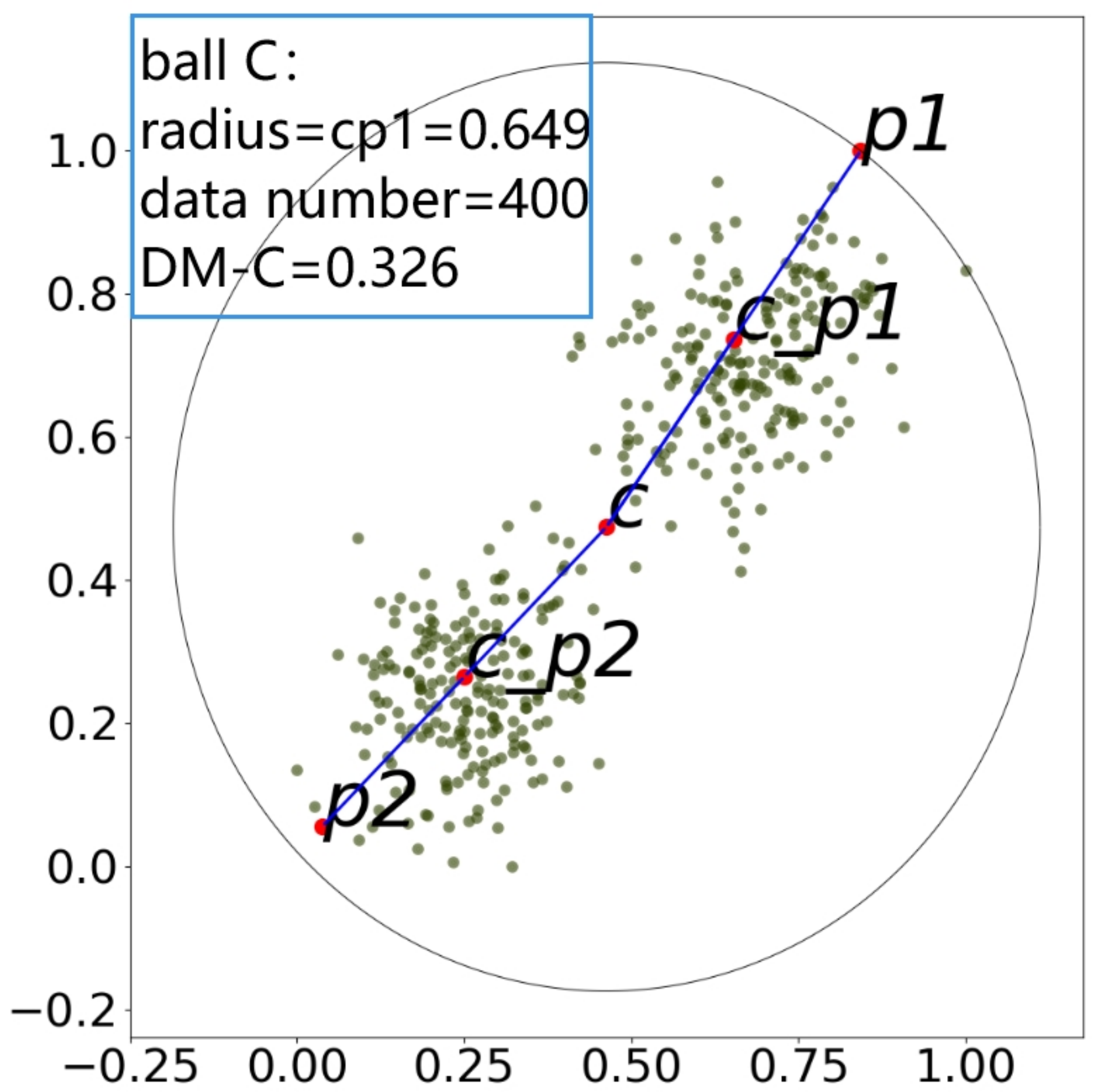}}
	\subfigure[]{
		\label{fig:subfig:Dataset4}
		\includegraphics[width=0.23\textwidth]{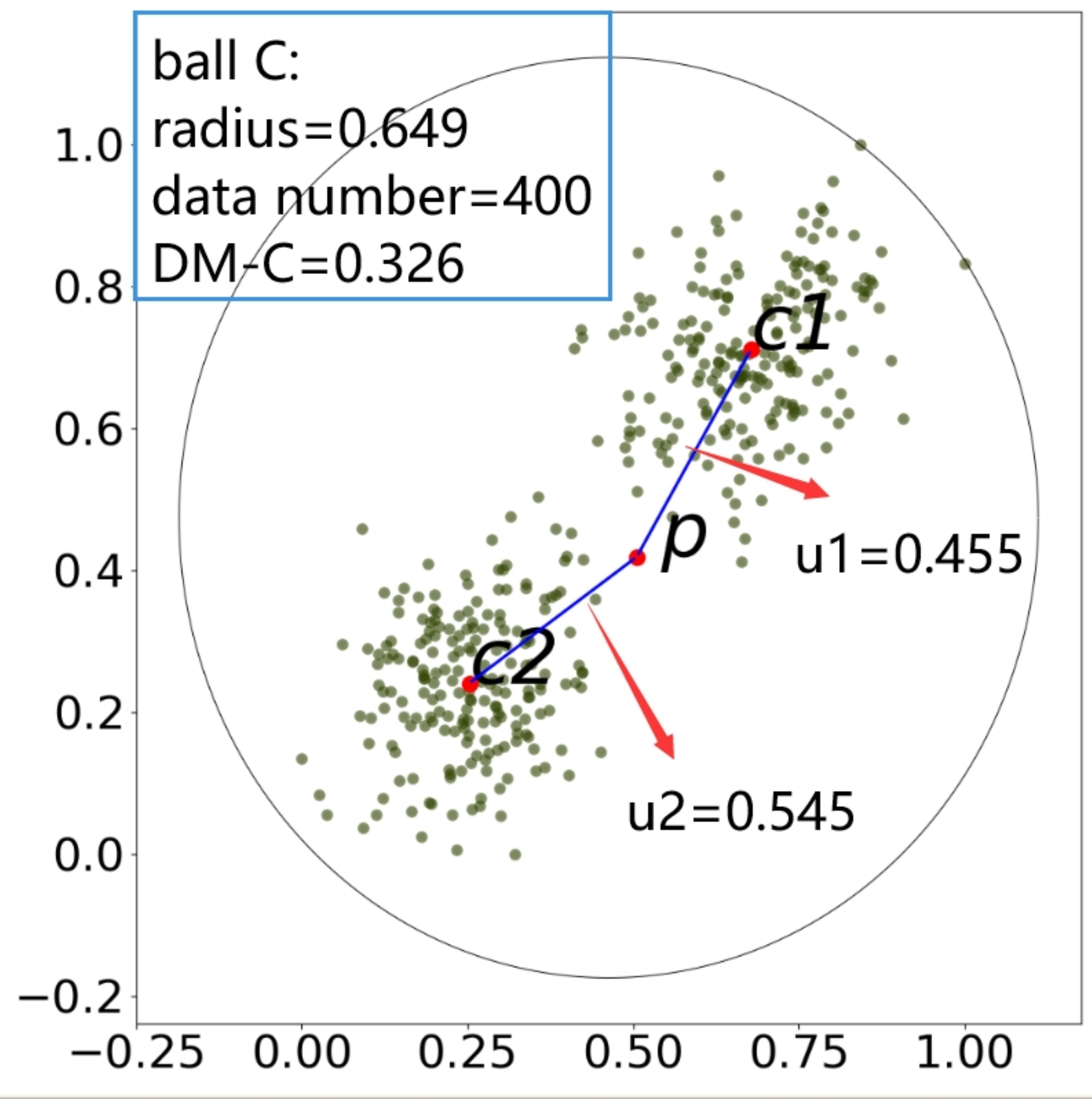}}
	\subfigure[]{
		\label{fig:subfig:Dataset4}
		\includegraphics[width=0.23\textwidth]{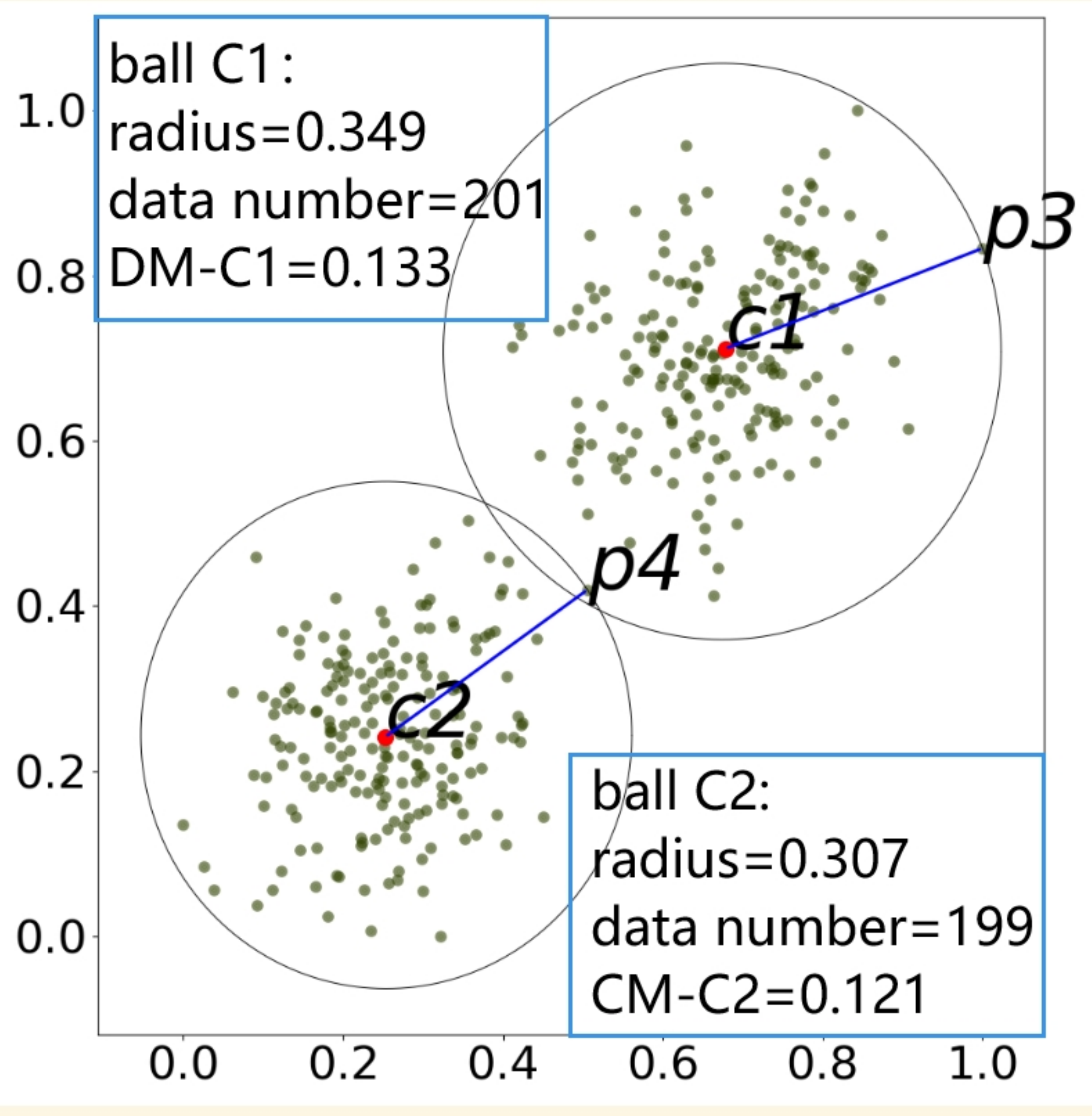}}
	\subfigure[]{
		\label{fig:subfig:Dataset4}
		\includegraphics[width=0.23\textwidth]{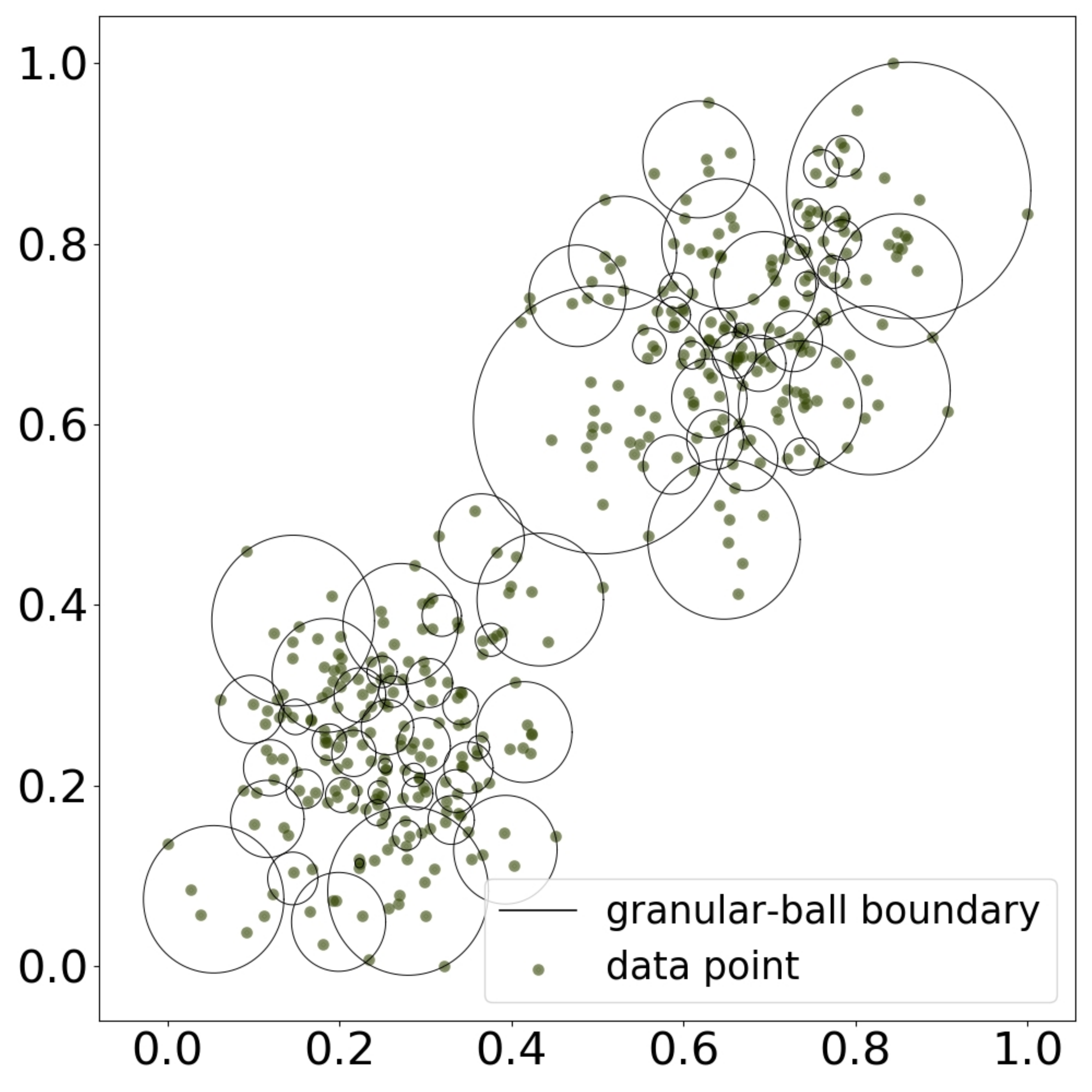}}
	\caption{The generation of fuzzy granular-balls.
		 a. p1 is the farthest point from the center c, and p2 is the farthest point from p1, here we select the mid-point c\_p1, c\_p2 of p1, p2 from c as the initial cluster center $V=[c\_p1,c\_p2]$.  
		 b. we calculate the membership U according to the initial V, and then iteratively update V and U according to the condition, finally we get ultimate cluster center $V=[c1,c2]$, if u2$ > $u1, then p belongs to cluster c2. 
		 c. if DM-C1$ < $DM-C and DM-C2$ < $DM-C, we split C1 as C1 and C2 according to  U  and V in b. 
		 d. Cycle through the process a to c until the generated quantity of fuzzy granular-balls remains the same ,so we get unnormalized fuzzy granular-balls. }.

	\label{fig:syn}
\end{figure*}

\subsection{Granular Computing}
Chen [29] pointed out that human cognition has the characteristics of “global precedence’ In 1982. It is different from the major existing artificial intelligence algorithms, which take the most fine-grained points as input. Wang [30] first introduced the large-scale cognitive rule into granularcomputing and proposed multi-granular cognitive computing. Xia and Wang [31]-[33] further used hyperspheres of different sizes to represent "grains", and proposed granular-ball computing, in which a large granular-ball represents coarse granu-larity, while a small granular-ball represents fine granularity.

Given a data set $  D  =  x_{i}(i = 1, 2,..., n) $, where n is the number of samples on D. Granular balls $ GB_{1}, GB_{2},..., GB_{m} $ are used to cover and represent the data set D. Suppose the number of samples in the $ j^{t h}  $ granular-ball $ GB_{i} $ is expressed as $ \left|G B_{j}\right| $, then its coverage degree can be expressed as $\sum_{i=1}^{m}\left(\left|G B_{j}\right|\right) / n  $. The basic model of granular-ball coverage can be expressed as

 \begin{equation}
 \label{equ:9}
\begin{array}{l}
\min \lambda_{1} * n / \sum_{j=1}^{m}\left(\left|G B_{j}\right|\right) / n+\lambda_{2} * m \text {, } \\
\text { s.t. quality }\left(G B_{j}\right) \geq T,
\end{array}
 \end{equation}

 where $ \lambda_{1} $ and $\lambda_{2} $ are the corresponding weight coefficients, and m the number of granular balls. When other factors remain unchanged, the higher the coverage, the less the sample information is lost, and the more the number of granular balls, the the characterization is more accurate. Therefore.the minimum number of granular-balls should be considered to obtain the maximum coverage degree when generating granular-balls. By adjusting the parameters Ai and $ \lambda_{2} $ the optimal granular-ball generation results can be obtained to minimize the value of the whole equation. In most cases, the two items in the objective function do not affect each otheland do not need trade off, so $ \lambda_{1} $ and $\lambda_{2} $ are set to 1 by default. Granular-ball computing can fit arbitrarily distributecdata [32],[34].

\section{The proposed method}

\subsection{Motivation }
There are a lot of fuzzy boundaries in the real world data, and hard division will lead to the uncertainty of the clustering results, which is also unreasonable. How to soft handle the clustering problem of data with fuzzy boundaries has great research value. As the most classic FCM algorithm in fuzzy clustering, many scholars have done a lot of research on its improvement. 
However, these FCM algorithms have considerable problems in noisy environments and are inaccurate for iterative calculations of clusters with a large number of different sample sizes.
 
The membership iteration of existing methods is mostly considered from the global perspective. This paper starts from the strategy of large-scale priority, and uses the granular-ball to perform fuzzy iteration on data. The membership of data only considers the two granular-ball, thus improving the efficiency of iteration. The fuzzy particle set formed can use more processing methods for different data scenarios, which enhances the practicability of fuzzy clustering algorithm.

\subsection{The process of generating fuzzy granular-balls }
Granular computing model uses granular-ball as "particle" to replace point input and a new granular classifier framework. By dividing the dataset into granules, a clear decision boundary can be obtained while the original distribution of data can be basically maintained. Granular computing is efficient, robust and scalable.In the existing granular computing process, the whole data is taken as the coarsest granularity, and then the feature of granular-ball is used to refine it from coarse granularity to fine granularity, so as to realize the scalable and robust computing process.
Based on this, here we give the following definitions:

\textbf{Definition 1} (Fuzzy Granular-Ball FGB, Fuzzy Granular-Balls sets $FGB_s$).
Based on the above granular computing model, combined with the existing classical fuzzy clustering method, starting from the traditional iteration of the membership of the data object, this paper optimizes and improves it to use the granular-ball to carry out fuzzy iteration of the data, and the membership of the data only considers the two granular-balls where it is located, such as this cycle, and finally generates a series of fuzzy granular-balls sets  through iteration. Our method is based on a basic assumption: data with similar distribution form a fuzzy granular-ball FGB shown in figure 1a, and adjacent fuzzy granular-balls  $FGB_s$ form a cluster .

\textbf{Definition 2} (Center c and Radius r). Given a data set $D\in R_d$, for each FGB, $ c_i $ is the center of gravity of all data points in FGB, and $ r_i $ is the maximum distance from all points $ p_i $ to c. The center c and radius r are defined as:
\begin{equation}
\label{equ:10}
c_{i}=\frac{1}{n} \sum_{i=1}^{n} p_{i}
\end{equation}

\begin{equation}
\label{equ:11}
r_{i}=\max \left(\left\|p_{i}-c_{i}\right\|\right)
\end{equation}

Here $ \|.\| $  denotes the 2-norm and n is the number of data
points in FGB.

\textbf{Definition 3} (Distributed Measure DM). $ DM_{i} $ is measured by calculating the ratio of data point $ n_{i} $ and radius $ s_{i} $ in FGB. here are definitions of them:

\begin{equation}
\label{equ:12}
s_{i}=\sum_{i=1}^{n_{i}}\left\|p_{i}-c i\right\|
\end{equation}

\begin{equation}
\label{equ:13}
D M_{i}=\frac{s_{i}}{n_{i}}
\end{equation}

\textbf{Definition 4} (Initial Cluster Center V). As shown in the figure. 1a : according to the center c of the given fuzzy granular-ball, we find the data point p1 that is farthest from c, and then find the data point p2 that is farthest from p1. Based on this, we select the mid-point c\_p1 of c and p1 and the mid-point c\_p2 of c and p2 as the initial cluster center, that is : $ V=[c\_p1, c\_p2] $.

\begin{equation}
\label{equ:14}
c_{-} p_{1}=\frac{1}{2}(c+p1)
\end{equation}

\begin{equation}
\label{equ:15}
c_{-} p_{1}=\frac{1}{2}(c+p2)
\end{equation}

 \begin{figure*}
	\centering
	\subfigure[unnormalized fuzzy granular-balls]{
		\label{fig:subfig:Dataset1}
		\includegraphics[width=0.23\textwidth]{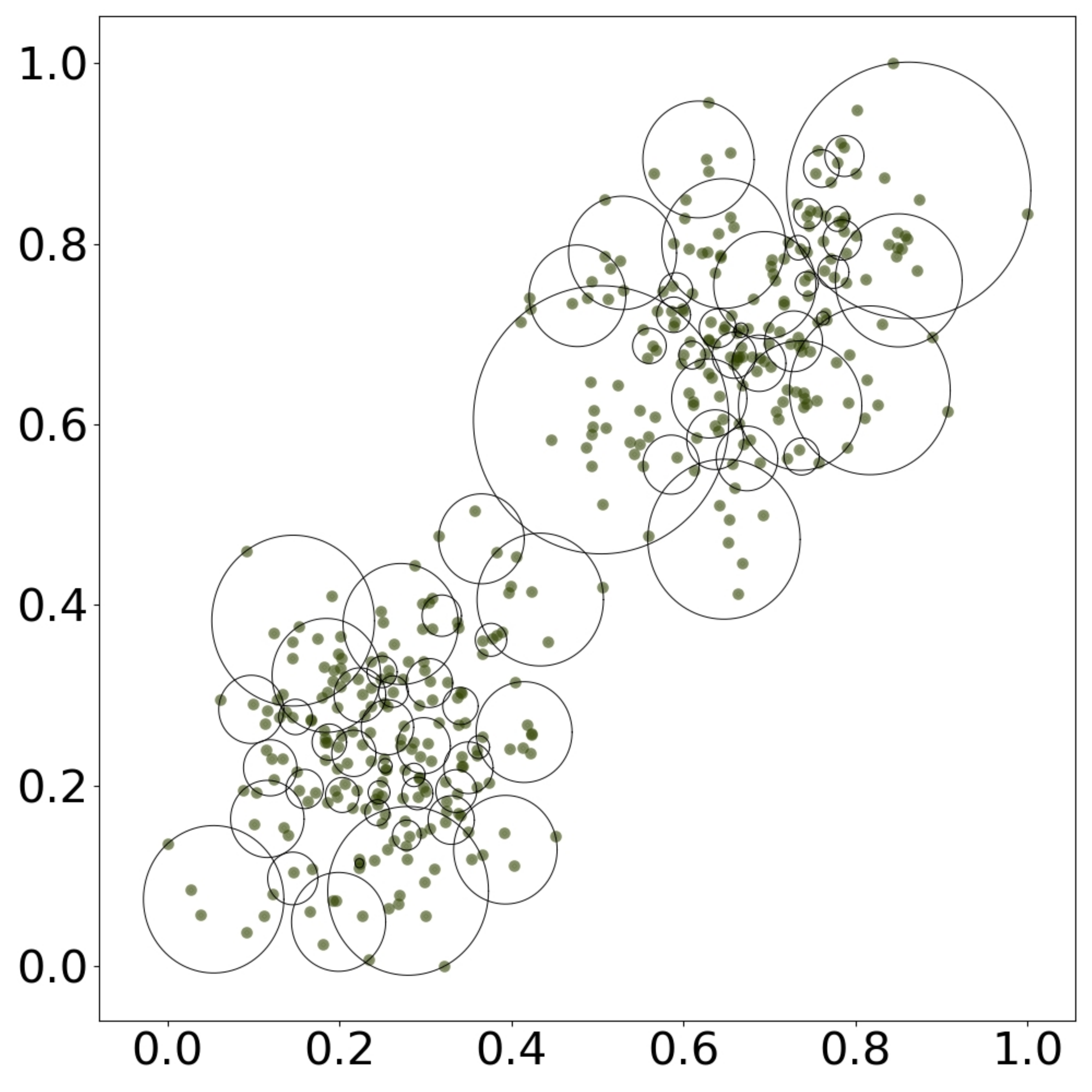}}
	\subfigure[before normalization]{
		\label{fig:subfig:Dataset1}
		\includegraphics[width=0.23\textwidth]{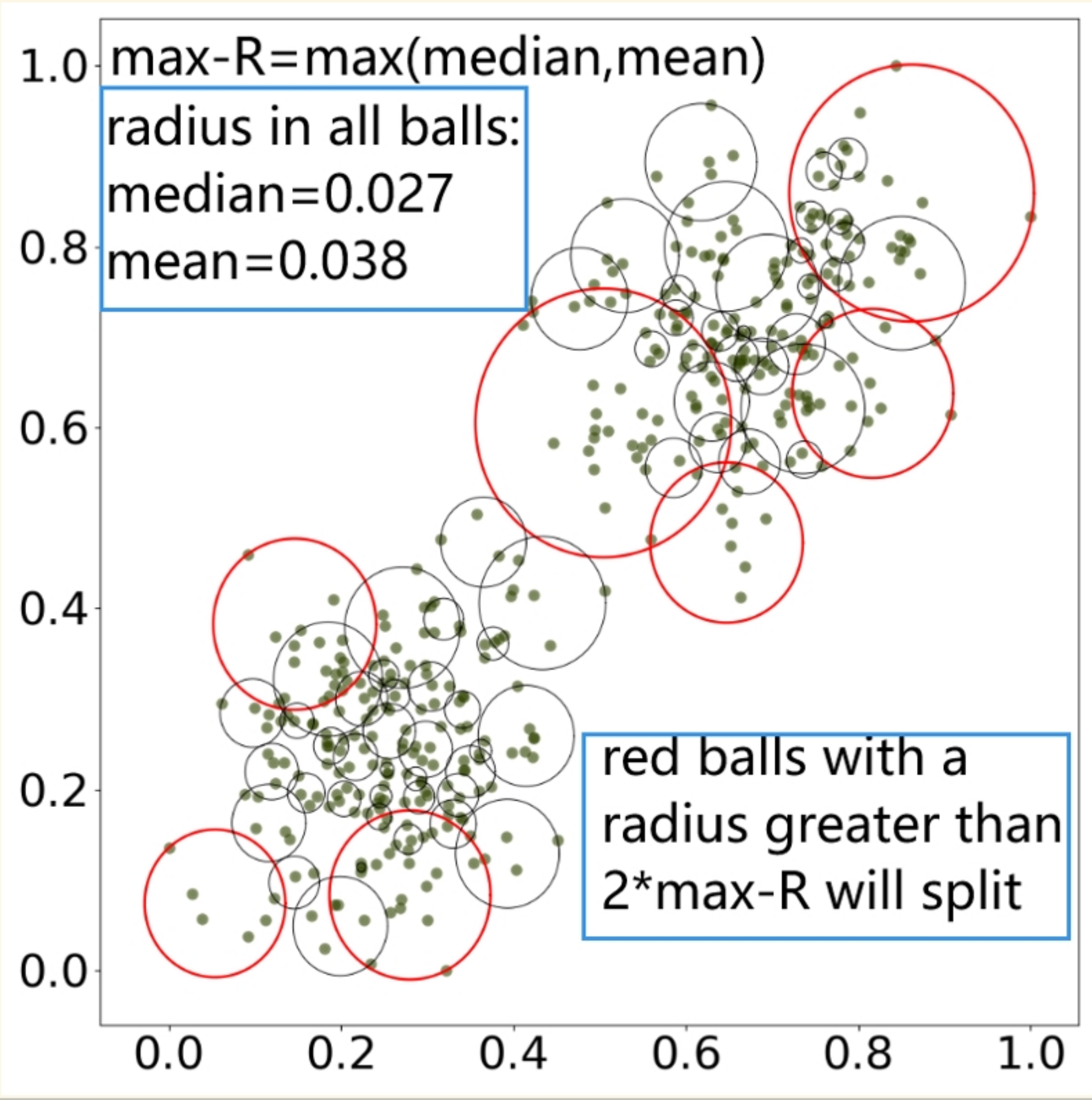}}
	\subfigure[after normalization]{
		\label{fig:subfig:Dataset2}
		\includegraphics[width=0.23\textwidth]{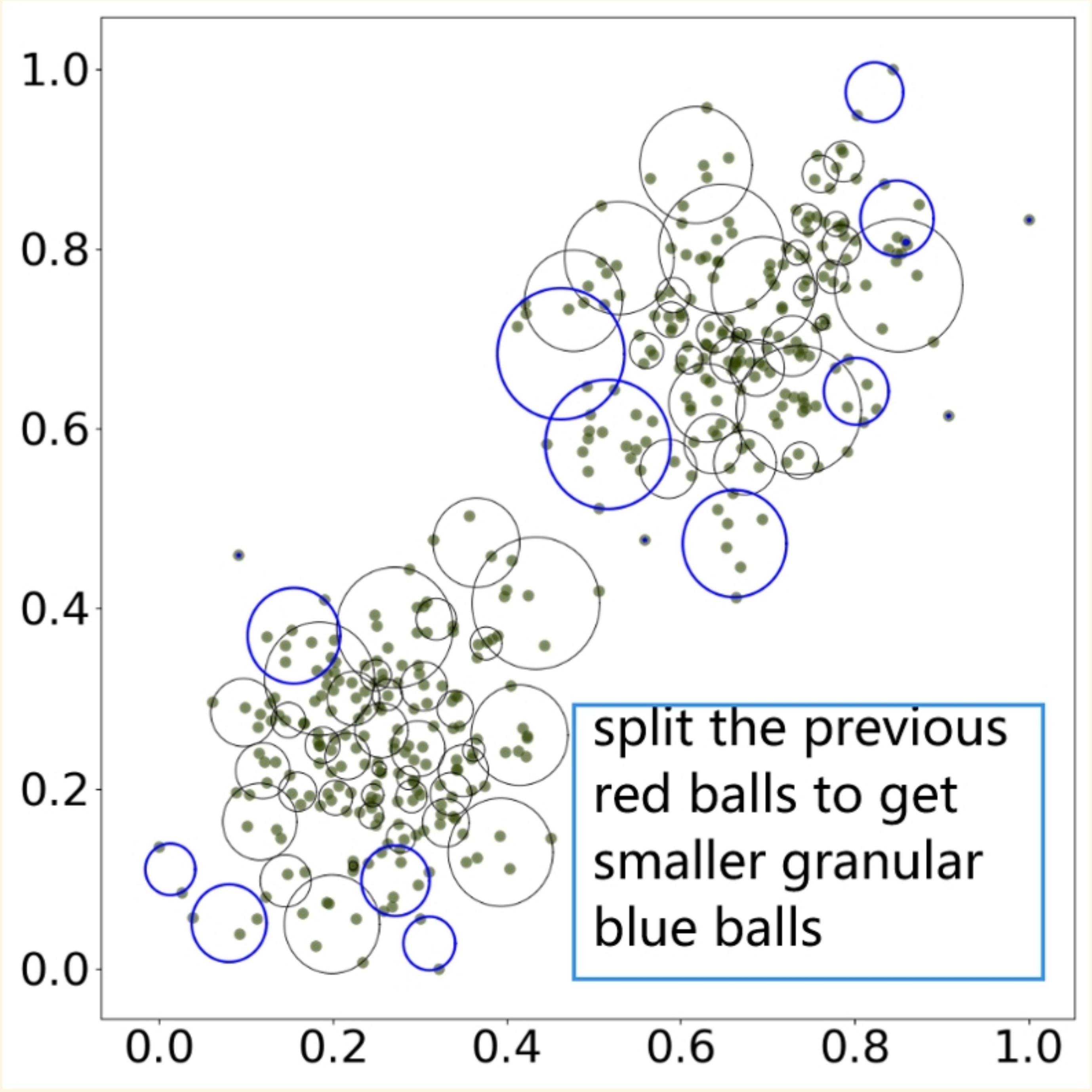}}
	\subfigure[final fuzzy granular-balls set]{
		\label{fig:subfig:Dataset3}
		\includegraphics[width=0.23\textwidth]{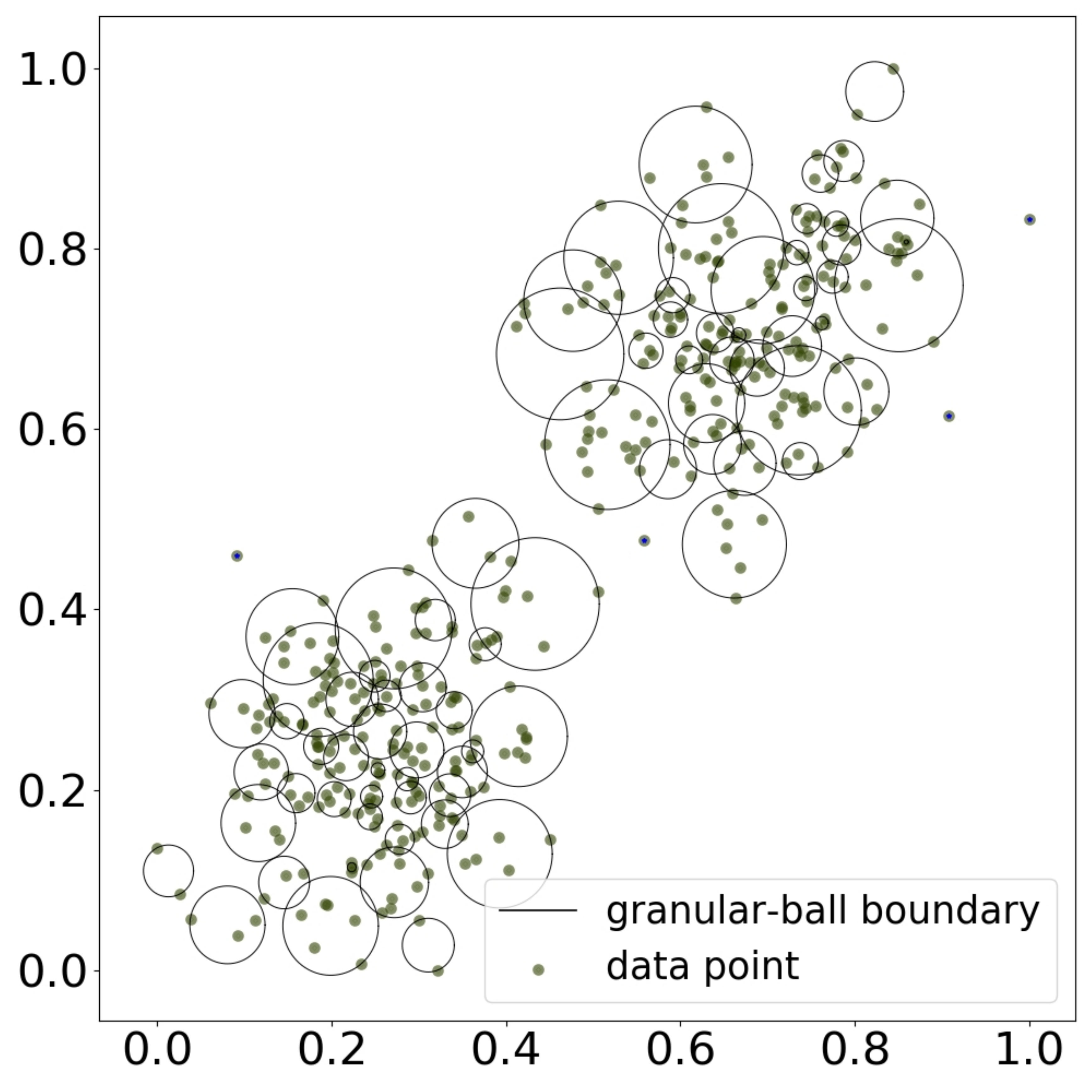}}
	\caption{The normalization process of fuzzy granular-balls.}
	\label{fig:syn}
\end{figure*}

\textbf{Definition 5} (Weighted DM value $DM_{weight}$). 
The fuzzy granular-ball is divided according to DM. As shown in the figure. 1a: We regard the entire dataset as a fuzzy granular-ball C; After determining the initial cluster center  $ V=[c\_p1, c\_p2] $, we use FCM algorithm to iteratively calculate the membership matrix U and update the cluster center  $ V=[c\_p1, c\_p2] $ in  figure. 1b. And then we split the fuzzy granular-ball into two sub fuzzy granular-ball  C1 and C2; In the whole process, we calculate DM-C, DM-C1 and DM-C2 values of C, C1 and C2; Finally, by comparing DM-C, DM-C1 and DM-C2, determine to finish splitting the ball in figure 1c. Finally we get a series of  unnormalized fuzzy granular-balls in figure 1d. In [x], DM-C1 and DM-C2 need to be smaller than DM-C to separate fuzzy granular-ball . However, when there is a lot of noise, this splitting rule will lead to a large number of fuzzy granular-ball  can not be split. Therefore, we use a weighted DM value for comparison in this paper, which can better adapt to noisy situations. To make the formula more concise and clear, we describe DM-C as $ D M_{C } $, DM-C1 as $ D M_{C 1} $, and DM-C2 as $ D M_{C 2} $. Therefore {$DM_{weight}$}     is defined as
follows:

\begin{equation}
\label{equ:16}
D M_{\text {weight }}=\frac{n_{C 1}}{n_{C}} 
D M_{C 1}+\frac{n_{C 2}}{n_{C}} D M_{C 2} \text {. }
\end{equation}

Here $  {n_{C}} $, ${n_{C1}}$  and $ {n_{C2}} $   respectively represent the number of data in the corresponding fuzzy granular-ball. If $DM_{weight}$ is smaller than $ D M_{C} $, fuzzy granular-ball C will split. Figure. 2a shows the final split result. However, in Figure. 2a, some fuzzy granular-balls with too large radius may still be affected by some boundary points or noise points, so segmentation is required; If {$r_k \ge 2\times  max(mean(r),median(r))$}    , you need to split $FGB_{K} $. Mean (r) and median (r) represent the average and median of all fuzzy granular-ball radius, respectively. After removing the fuzzy granular-ball with too large radius, the splitting process is completed, and the results are shown in Fig. 2b. Based on the above description, a fuzzy granular-ball generation algorithm is designed and shown in Algorithm 1.

\begin{algorithm}
	\renewcommand{\algorithmicrequire}{\textbf{Input:}}
	\renewcommand{\algorithmicensure}{\textbf{Output:}}
	
	\caption{Generation of fuzzy granular-balls}
	\label{alg:1}
	
	\begin{algorithmic}[1]
		\REQUIRE $D$: the dataset  
		\ENSURE $FGB_s$ $sets$: the fuzzy granular-balls sets 
		\renewcommand{\algorithmicensure}{\textbf{Initialize:}}
		\ENSURE  $n=2$, $m=2$,$iter=0$, $max\_count=100$\;
		
		\WHILE {true}
		\STATE initialize  $FGB_{s}=\emptyset$ 
		\STATE initialize  $V=[c\_p1,c\_p2]$
		according to Eq.14,Eq.15;
		\STATE calculate U according to Eq.8;
		\WHILE {$iter<max\_count$}
		
		\STATE calculate V according to Eq.9;
		\STATE update U according to Eq.8;
		\STATE iter=iter+1;
		\ENDWHILE
	
		\STATE pre-split D as $FGB_{k} $ according to the U;
		\STATE calculate DM of D,$DM_{weight}$ according to Eq.16;
		\IF {DM $\le$ $DM_{weight}$}          
		\STATE remain the $FGB_{k}$ and add it to  set $FGB_{s}$ ;
		\ENDIF
		\IF {the number of   $FGB_{k}$ is not changing}          
		\STATE break;
		\ENDIF
		\STATE Take the data $D_{k} $ in $C_{k} $ as the next input, $ D=D_{k} $
		\ENDWHILE

		\FOR{each $FGB_{k} \in FGB_{s} $}
		\STATE calculate mean(r),median(r),
		\IF {$r_k \ge 2\times  max(mean(r),median(r))$}          
		\STATE Split $FGB_k$;
		\ENDIF
			\IF {the number of   $FGB_{k}$ is not changing}          
		\STATE break;
		\ENDIF
		
		\ENDFOR
		\STATE return $FGB_s$
		
	\end{algorithmic}  

\end{algorithm}

\subsection{Connect fuzzy granular-balls to form cluster results}
\textbf{Definition 6} (connecting Fuzzy Granular-Balls sets by K-means  FGB\_K-means). 
 according to the algorithm 1, we get final fuzzy granular-balls set, now we need to connect these fuzzy granular-balls to form final cluster results. So here we provide two ways to connect, one is to use these fuzzy granular-balls as the input of K-means algorithm to complete clustering, which is defined as FGB\_K-means. So this algorithm is designed and shown in Algorithm 2.
 
 \begin{algorithm}
 	\renewcommand{\algorithmicrequire}{\textbf{Input:}}
 	\renewcommand{\algorithmicensure}{\textbf{Output:}}
 	
 	\caption{ FGB\_K-means}
 	\label{alg:1}
 	
 	\begin{algorithmic}[1]
 		\REQUIRE $FGB_s$ $sets$: the fuzzy granular-balls sets, the num of clusters: K
 		\ENSURE The Clutering result 

 		\STATE initialize $center\_list=\emptyset$;
 		\FOR{each $FGB_{i} \in D$}
 		\STATE calculate the center of $FGB_{i}$,
 		\STATE add  the center of $FGB_{i}$ to center\_list
 		
 		\ENDFOR
 		\STATE consider center\_list as dataset D, and the dataset D and K are  used as input of K-means to connect $FGB_s$ to final clustering results
 		\STATE return clustering results
 		
 	\end{algorithmic}  
 	
 \end{algorithm}
\textbf{Definition 7} (connecting Fuzzy Granular-Balls sets by overlap  relationship FGB\_overlap). 
 the other is our proposed method, which is defined FGB\_overlap, it is based on one of the connectivity relationship, which is formed on adjacent rules. The adjacent rules are described as follows:
 \begin{figure*}
 	\centering
 	\includegraphics[width=0.9\linewidth]{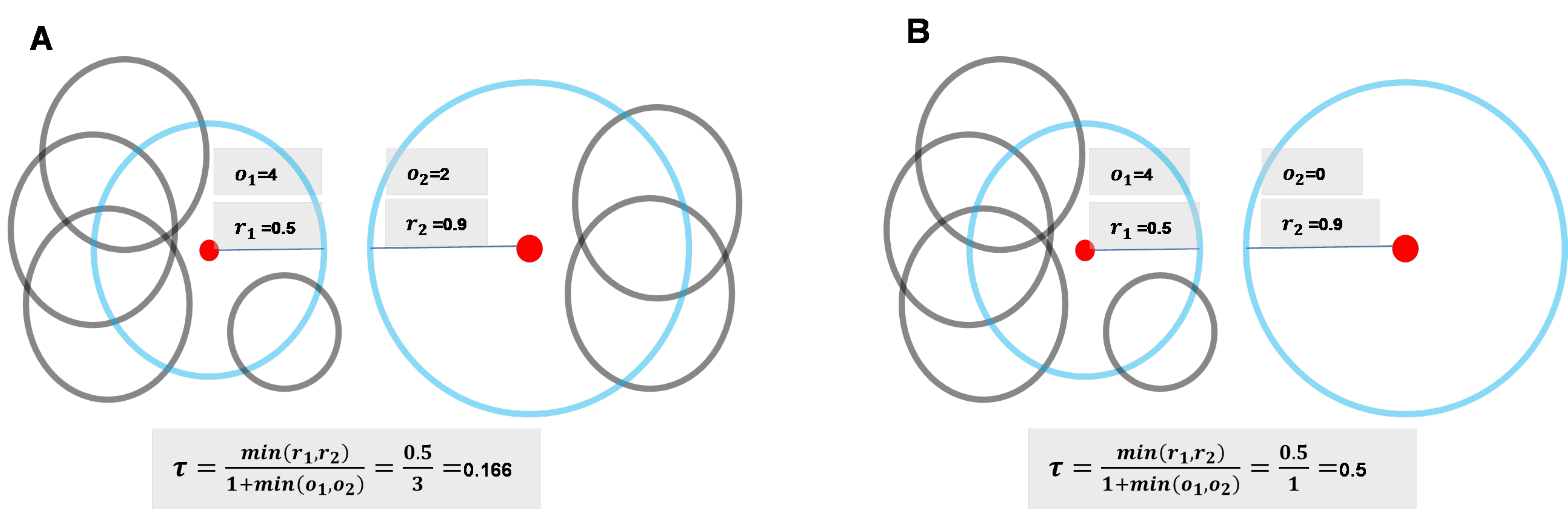}
 	\caption[acc]{$ o_1 $ and $ o_2\ $ are the cumulative values of the overlapping times of the neighbor fuzzy granular-ball as the adjustment coefficients of ?, which can dynamically adjust the criterion that meets the adjacent rules. To prevent the wrong differentiation of the fuzzy granular-balls, the larger the cumulative value is, the stricter the standard that meets the adjacent rules. The $ \tau $ in A is smaller than the $ \tau  $ in B.}
 	\label{fig:mr-acc}
 \end{figure*}
 \begin{equation}
 \label{equ:17}
\left\|c_{i}-c_{j}\right\|-\left(r_{i}+r_{j}\right)<\tau_{i j}
 \end{equation}
 
 \begin{equation}
\label{equ:18}
\tau_{i j}=\frac{\min \left(r_{i}, r_{j}\right)}{1+\min \left(o_{i,} o_{j}\right)}
\end{equation}

where $ o_i $ represents the cumulative values of the overlapping times between $ {FHB}_i $ and its neighbor fuzzy granular-balls, and $ o_j $ represents that between $ {FHB}_j $ and its neighbor fuzzy granular-balls; $ \tau_{ij} $ denotes the adjustment coefficients. The formula shows that if the gap between $ {FHB}_i $ and $ {FHB}_j $ is less than  $ \tau_{ij} $, they meet the adjacent rules. $ \tau_{ij} $ can dynamically adjust the criterion that meets the adjacent rules. As shown in Fig. 3, to prevent the wrong differentiation of the granular-balls, the larger the cumulative value is, the stricter the standard that meets the adjacent rules. If two granular-balls meet adjacent rules, they belong to the same cluster. Based on the above description, a fuzzy granular-ball connection algorithm is designed and shown in Algorithm 3.

\begin{algorithm}
	\renewcommand{\algorithmicrequire}{\textbf{Input:}}
	\renewcommand{\algorithmicensure}{\textbf{Output:}}
	
	\caption{FGB\_overlap}
	\label{alg:1}
	
	\begin{algorithmic}[1]
		\REQUIRE $FGB_s$ $sets$: the fuzzy granular-balls sets
		\ENSURE The Clutering result 
	
		\FOR{each $FGB_{i} \in FGB_s$}
		\FOR{each $FGB_{j}(i \neq j) \in FGB_s$}
		\STATE  Calculate the overlapping times  $o_{i}$ of $FGB_{i}$ and $FGB_{j}$  
		\ENDFOR
		
		\ENDFOR
		\FOR{each $FGB_{i} \in FGB_s$}
				\FOR{each $FGB_{j}(i \neq j) \in FGB_s$}
			\IF {$FGB_{i}$ and $FGB_{j}$  meet the overlap rules according to Eq.17, E1.18}          
			\STATE connect them and set with the same label;
			\ENDIF
			
			\ENDFOR
		\ENDFOR
		
		\FOR{each $FGB_{i} \in FGB_s$}
		\FOR{each $FGB_{j}(i \neq j) \in FGB_s$}
		\IF  {$o_{i}=0   $  }    
		\STATE connect $FGB_{i}$ or $FGB_{j}$ with the nearest FGB and set with the same label
		\ENDIF
		
		\ENDFOR
		\ENDFOR
		\STATE return clustering results
		
	\end{algorithmic}  
	
\end{algorithm}

\section{Experiments}

in this section, all of related methods are evaluated on synthetic and real datasets. Since our method is based on the application and improvement of FCM, we have done comparative experiments on FCM, ECM-NSGA-II, ECM-MOEA/D, MFS-FCM, DI-FCM, RI-FCM, FDPC, and the K-means algorithm is also compared as a reference. The detailed configuration of experiment hard-ware is as follows: Intel Core i7-10700 CPU @ 2.90GHz, and 16G RAM. Programming environment is PyCharm.

Three criteria were used to evaluate clustering performance on the labeled dataset: ACC [35] and NMI [36],[37], ARI [38]. When calculating ACC, a label needs to be assigned to each data. The process of assigning labels is random, which will lead to mismatches with the real labels and make the calculation of ACC wrong. Therefore, a mapping function needs to be constructed to match the labels obtained by clustering with the real labels one by one. Based on this mapping function,
the formula for calculating ACC is:

\begin{figure*}
	\centering
	\subfigure[GroundTruth]{
		\label{fig:subfig:Dataset3}
		\includegraphics[width=0.23\textwidth]{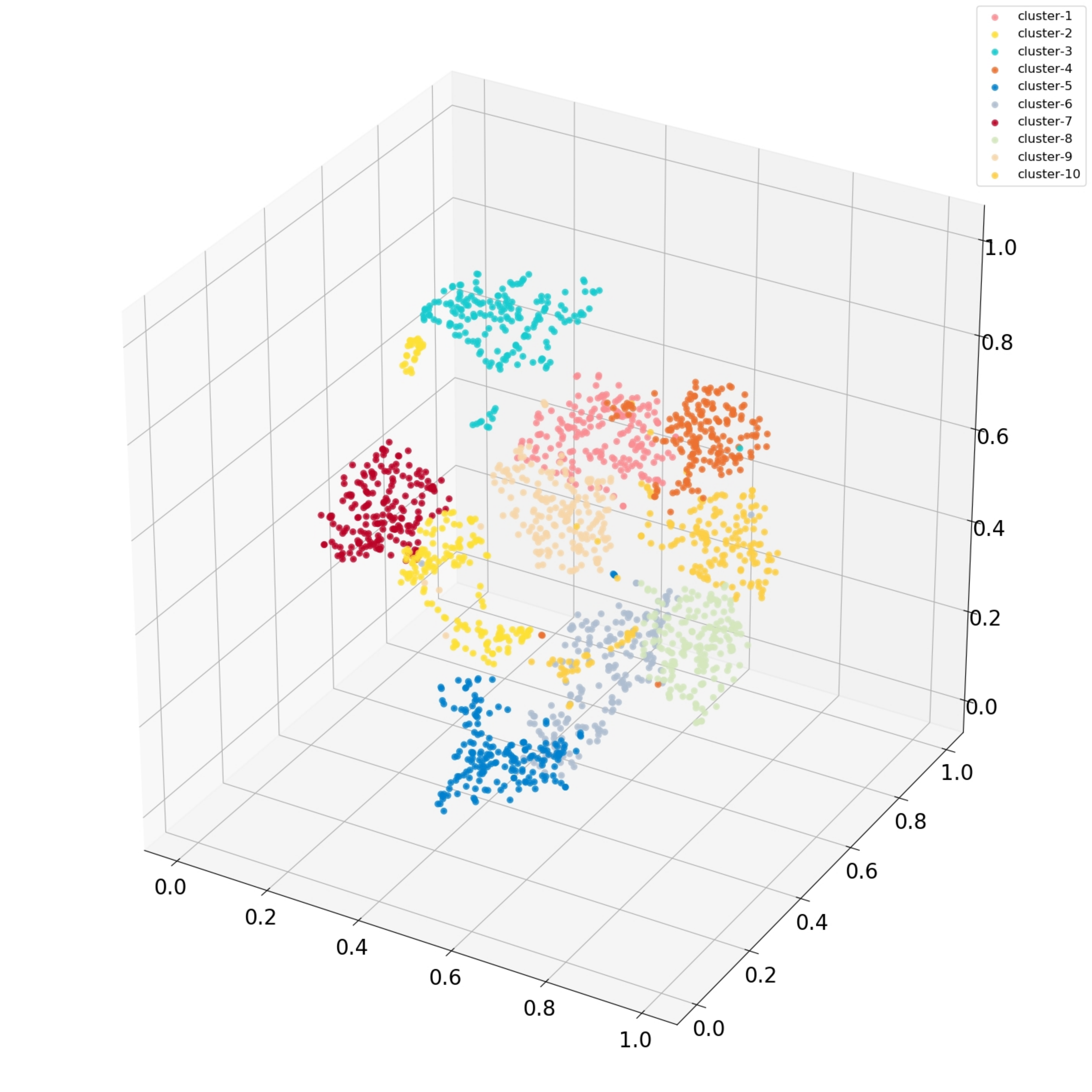}}
	\subfigure[generated fuzzy granular-balls]{
		\label{fig:subfig:Dataset1}
		\includegraphics[width=0.23\textwidth]{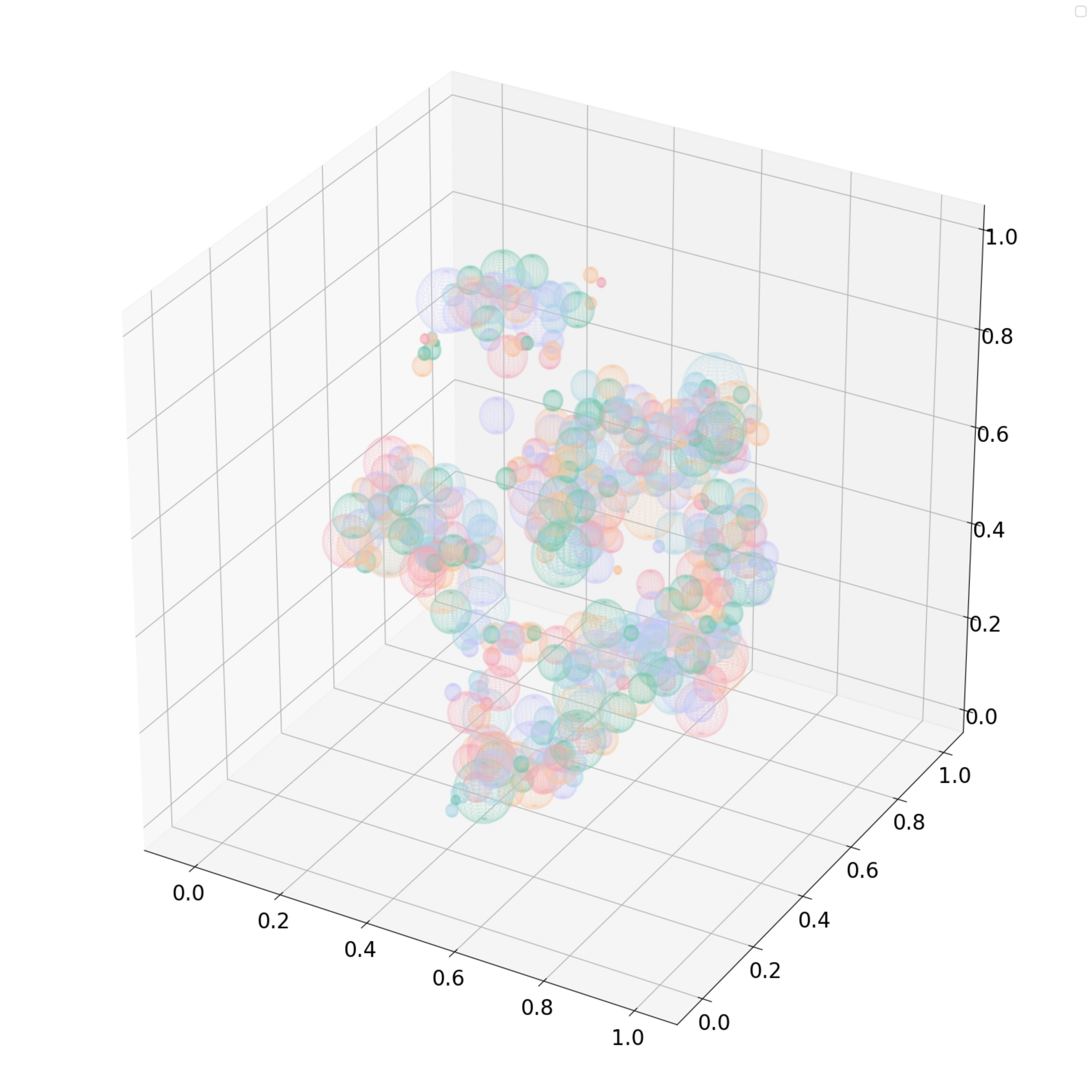}}
	\subfigure[FGB\_overlap]{
		\label{fig:subfig:Dataset1}
		\includegraphics[width=0.23\textwidth]{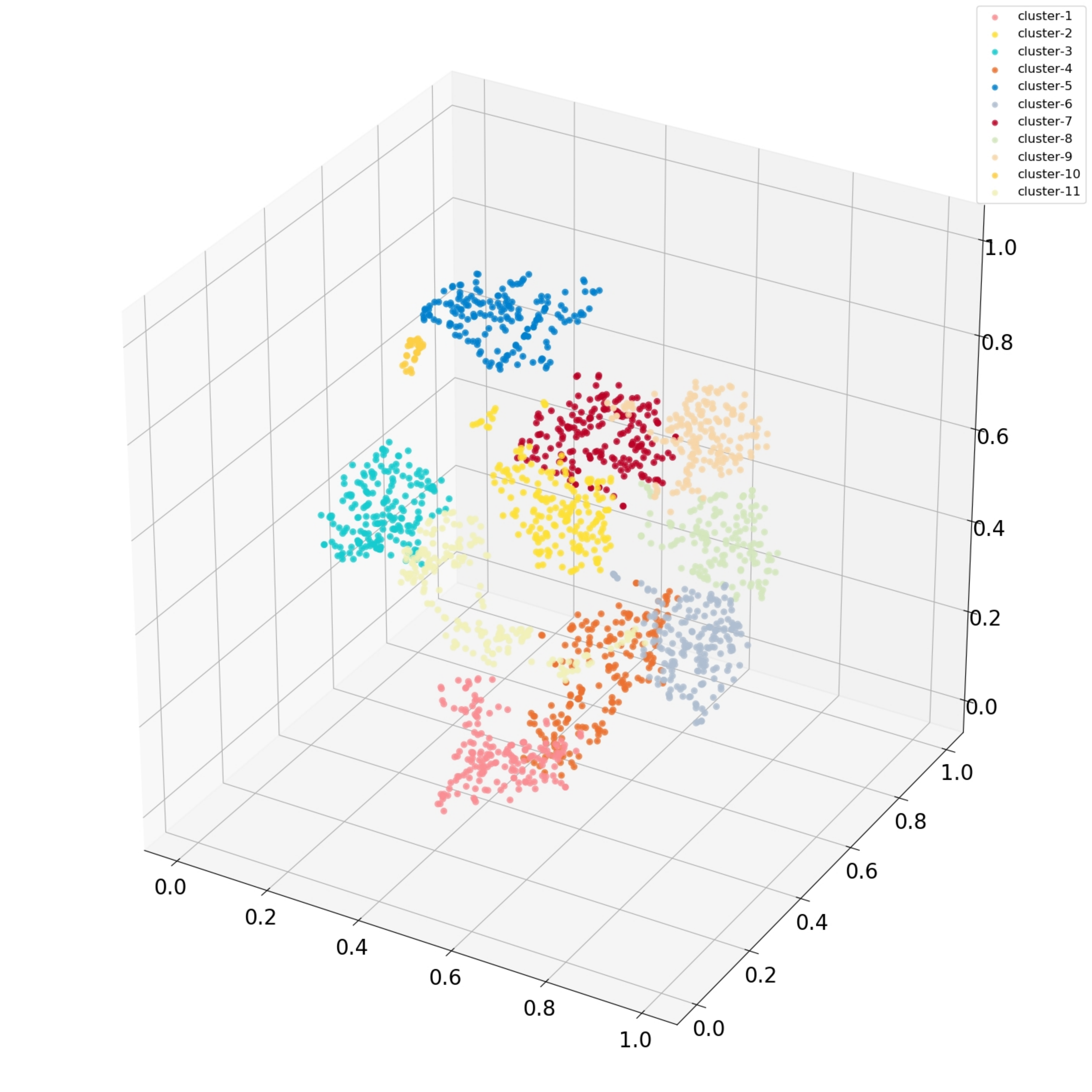}}
	\subfigure[FGB\_K-means]{
		\label{fig:subfig:Dataset2}
		\includegraphics[width=0.23\textwidth]{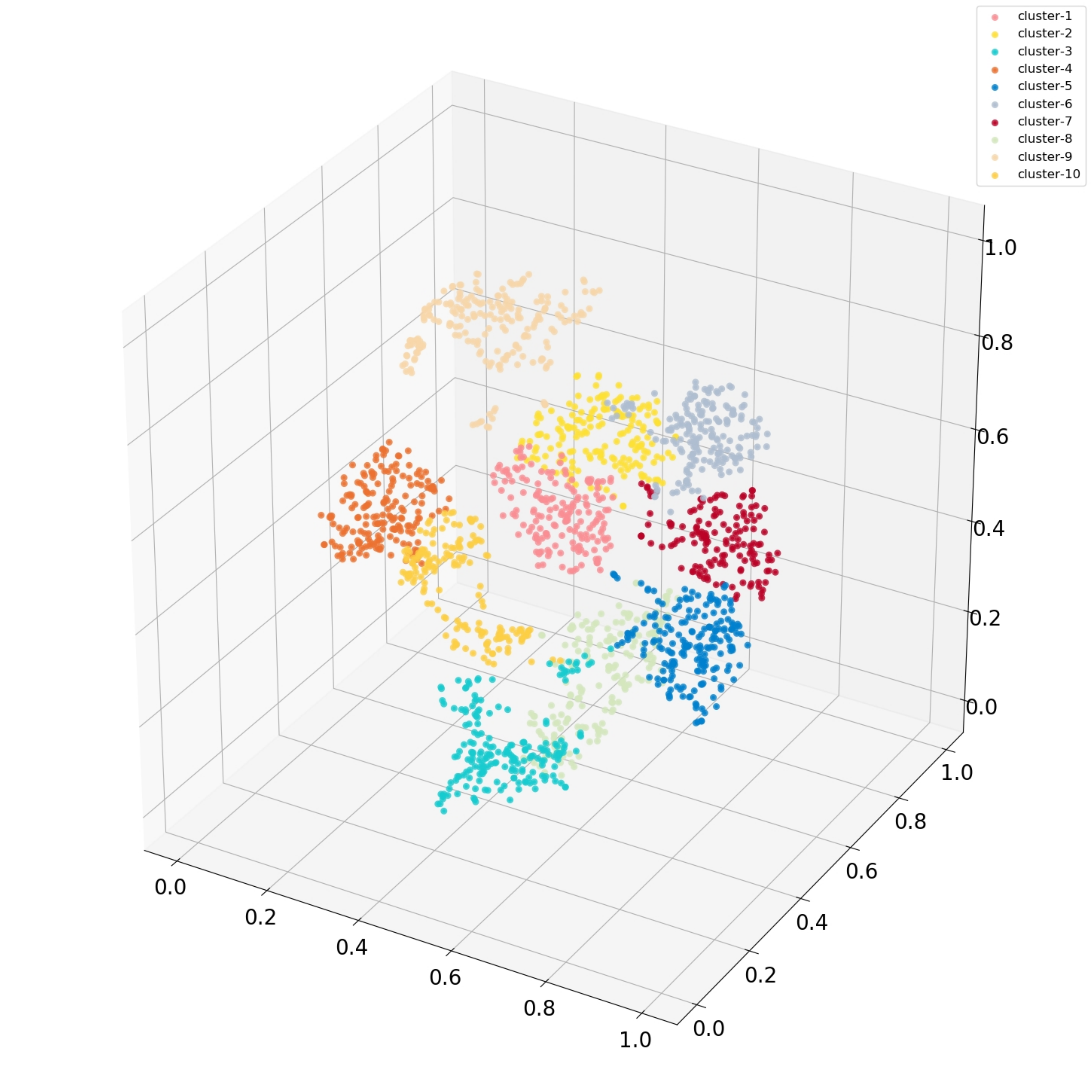}}
	
	\caption{The clustering result of our methods on mnist dataset.}
	\label{fig:syn}
\end{figure*}

 \begin{equation}
\label{equ:19}
A C C=\frac{1}{n} \sum_{i=1}^{n} \delta\left(l_{i}, \operatorname{map}\left(c_{i}\right)\right)
\end{equation}

where $  l_i $ is the real label, $ c_i $ is the label obtained aftef clustering, and
$ \delta(x, y)=\left\{\begin{array}{ll}
	1 & x=y \\
	0 & x \neq y
\end{array}\right. $
is a discriminate
function.

Another performance evaluation metric is normalized mu-
tual information(NMI). Unlike ACC, NMI is not affected by
the arrangement of cluster labels. The formula for calculating
NMI is:
 \begin{equation}
\label{equ:20}
N M I(L, C)=2 \frac{M I(L, C)}{H(L)+H(C)}
\end{equation}
where L is the real label sequence, C is the clustered label
sequence, M I(L, C) refers to the mutual information between
L and C, H()representing the information entropy, and the
calculation formula of NMI can also be expressed as:

 \begin{equation}
\label{equ:21}
N M I(L, C)=\frac{\sum_{i=1}^{k} \sum_{j=1}^{k} n_{i, j} \log \left(n \cdot n_{i, j}\right)}{\sqrt{\left(\sum_{i} n_{i} \log \frac{n_{i}}{n}\right)\left(\sum_{j} n_{j} \log \frac{n_{j}}{n}\right)}}
\end{equation}
Finally, the last evaluation criteria ARI is
defined as follows:
\begin{tiny}
 \begin{equation}
\label{equ:21}
ARI =\frac{\sum_{i j}\left(\begin{array}{c}
	n_{i j} \\
	2
	\end{array}\right)-\left[\sum_{i}\left(\begin{array}{c}
	A_{i} \\
	2
	\end{array}\right)-\sum_{j}\left(\begin{array}{c}
	B_{j} \\
	2
	\end{array}\right)\right] /\left(\begin{array}{c}
	n \\
	2
	\end{array}\right)}{\frac{1}{2}\left[\sum_{i}\left(\begin{array}{c}
	A_{i} \\
	2
	\end{array}\right)+\sum_{j}\left(\begin{array}{c}
	B_{j} \\
	2
	\end{array}\right)\right]-\left[\sum_{i}\left(\begin{array}{c}
	A_{i} \\
	2
	\end{array}\right) \sum_{j}\left(\begin{array}{c}
	B_{j} \\
	2
	\end{array}\right)\right] /\left(\begin{array}{c}
	n \\
	2
	\end{array}\right)}
\end{equation}
\end{tiny}

%

\begin{table}[htbp]
	\centering
	\caption{The detailed information of real datasets}
	\begin{tabular}{cccc}
		\hline
		Dataset     & Instances & Dimensions & Clusters \\ \hline
		cell        & 8681      & 22         & 21       \\
		mnist       & 1797      & 3          & 10       \\
		Iris        & 150       & 4          & 3        \\
		Soybean     & 47        & 35         & 4        \\
		dermatology & 358       & 34         & 6        \\
		banknote    & 1372      & 4          & 2        \\
		ukm         & 258       & 5          & 4        \\
		segment     & 2310      & 19         & 7        \\ \hline
	\end{tabular}
	\label{tab:real}%
\end{table}%

\begin{figure*}
	\centering
	\subfigure[K-means]{
		\label{fig:subfig:Dataset3}
		\includegraphics[width=0.23\textwidth]{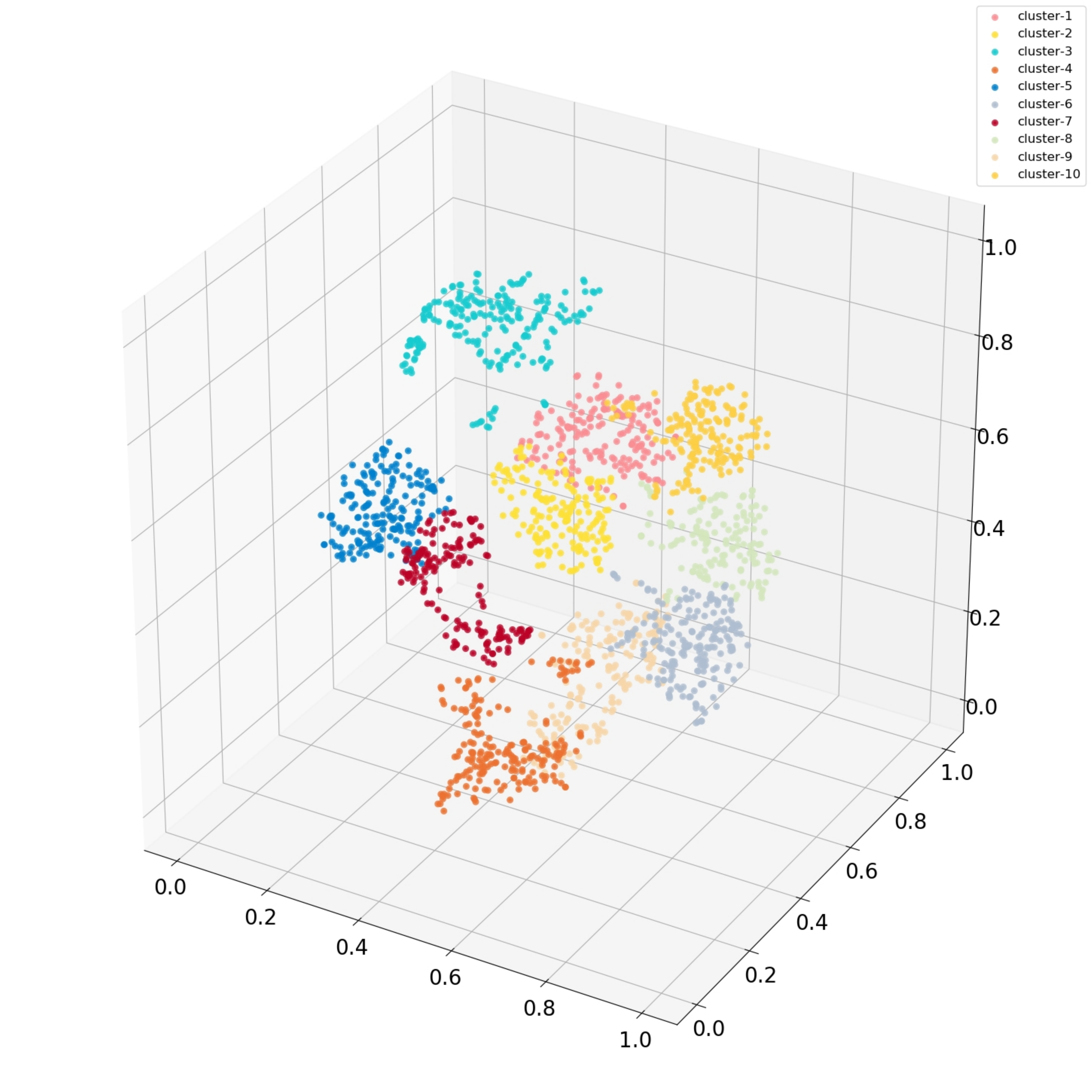}}
	\subfigure[FCM]{
		\label{fig:subfig:Dataset3}
		\includegraphics[width=0.23\textwidth]{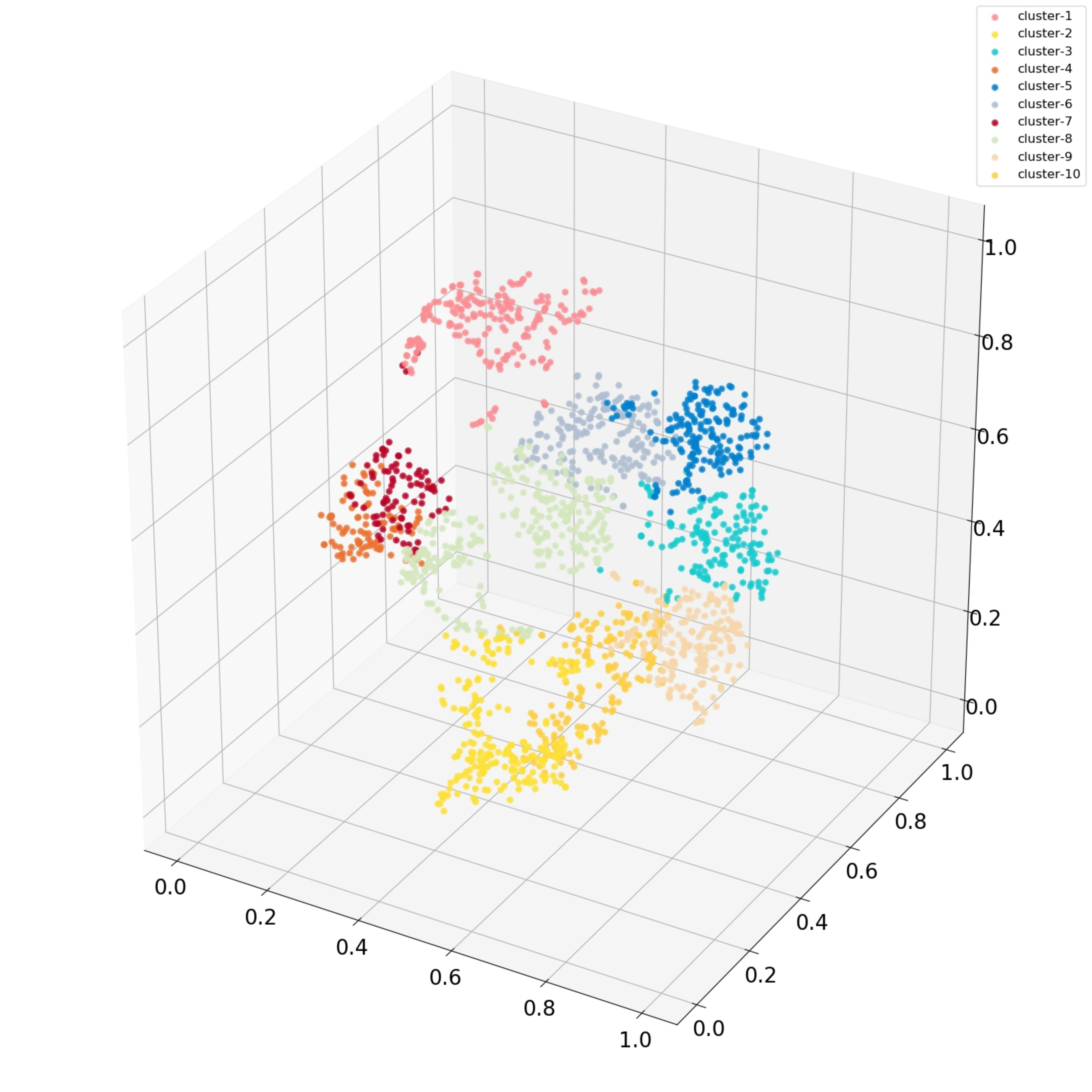}}
	\subfigure[ECM-NSGA-II 
	]{
		\label{fig:subfig:Dataset3}
		\includegraphics[width=0.23\textwidth]{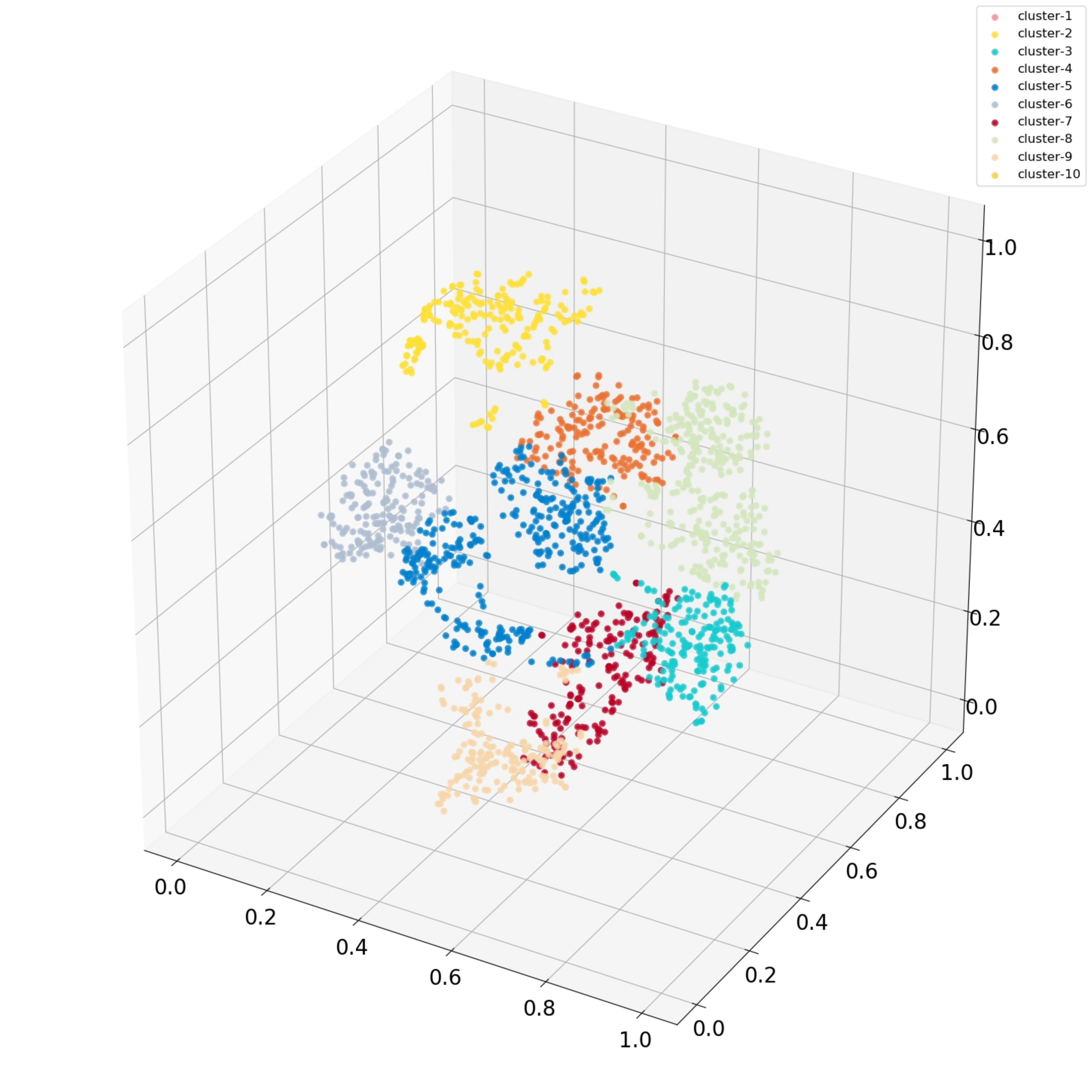}}
	\subfigure[ECM-MOEA/D
	]{
		\label{fig:subfig:Dataset3}
		\includegraphics[width=0.23\textwidth]{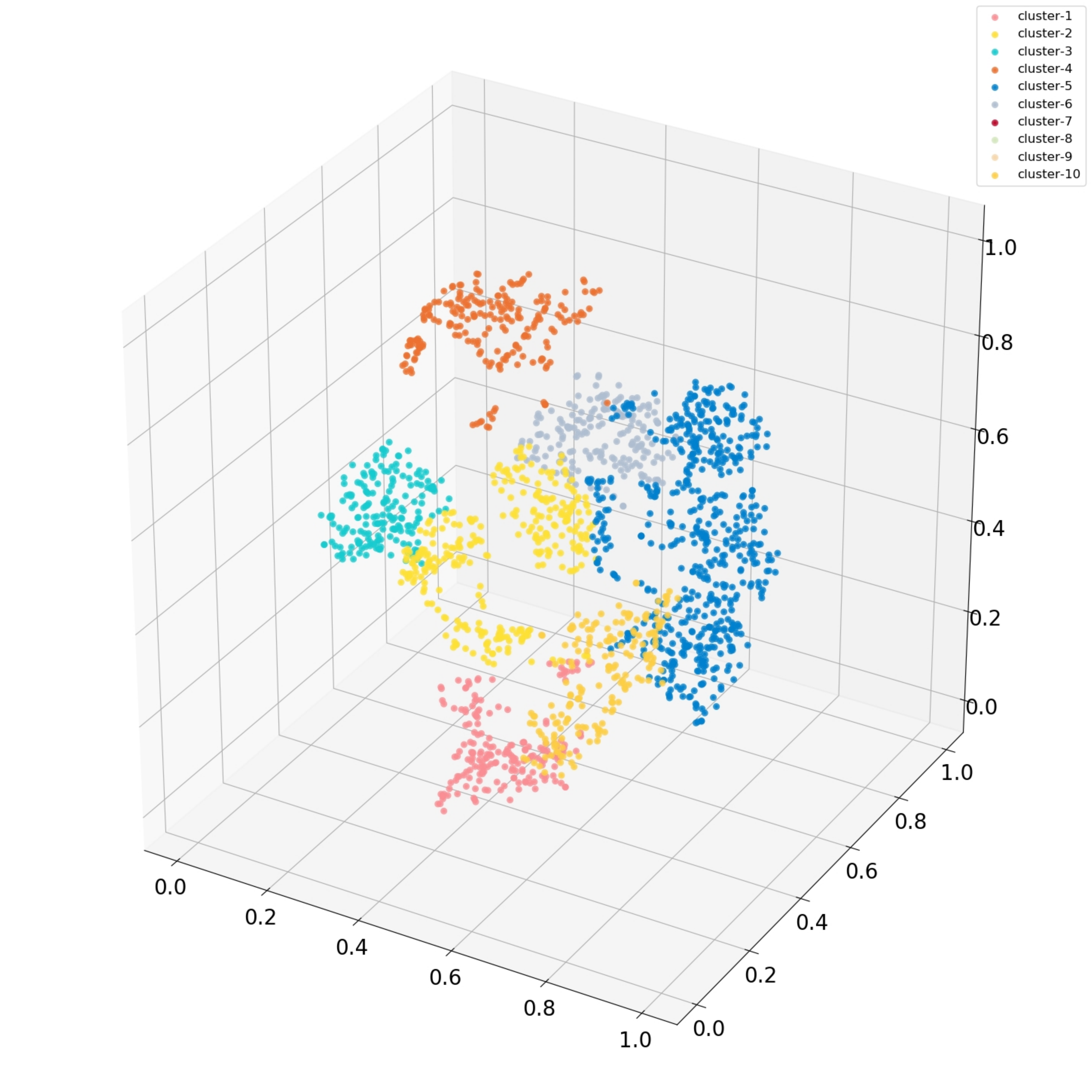}}
	
	\subfigure[MFS-FCM
	]{
		\label{fig:subfig:Dataset3}
		\includegraphics[width=0.23\textwidth]{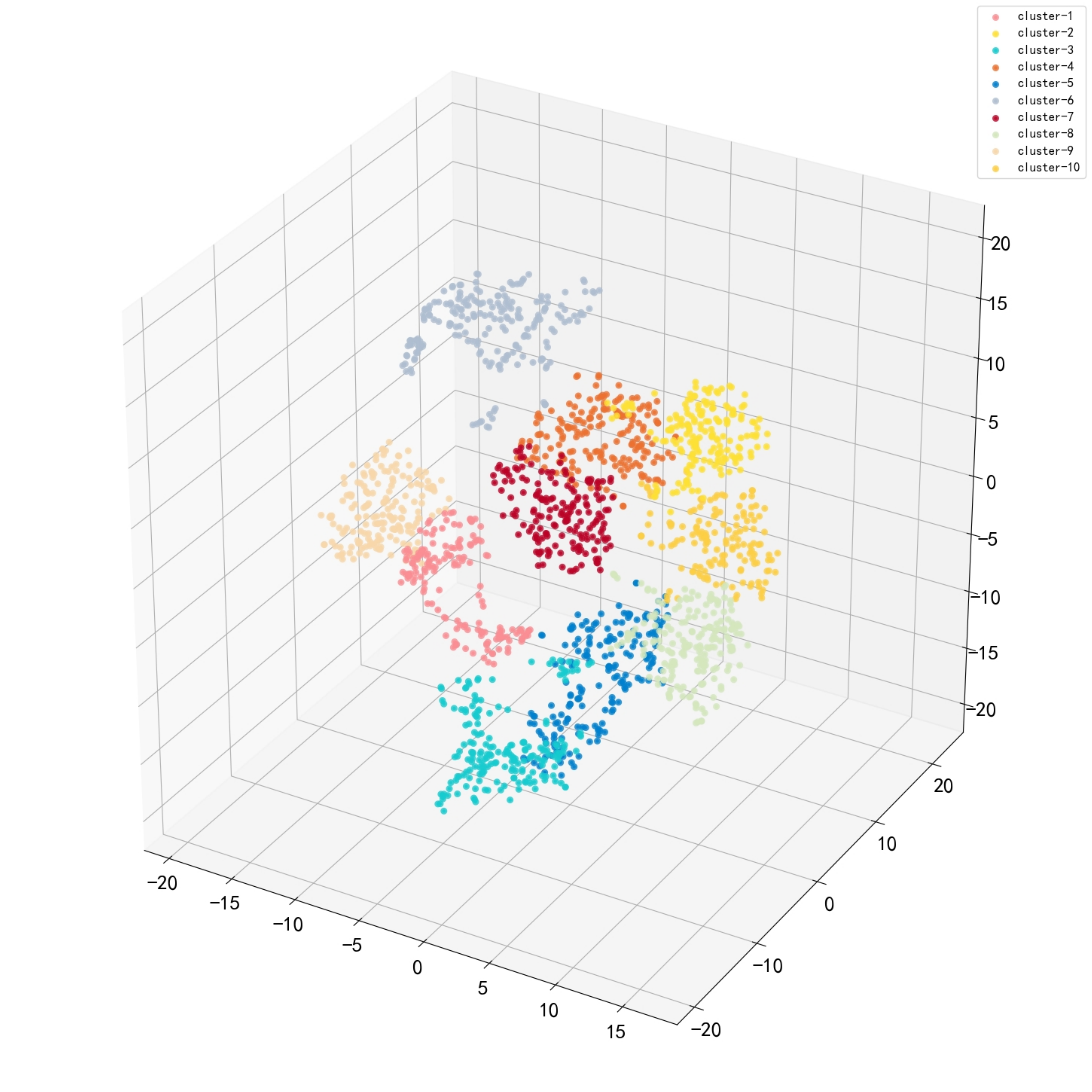}}
	\subfigure[DI-FCM
	]{
		\label{fig:subfig:Dataset3}
		\includegraphics[width=0.23\textwidth]{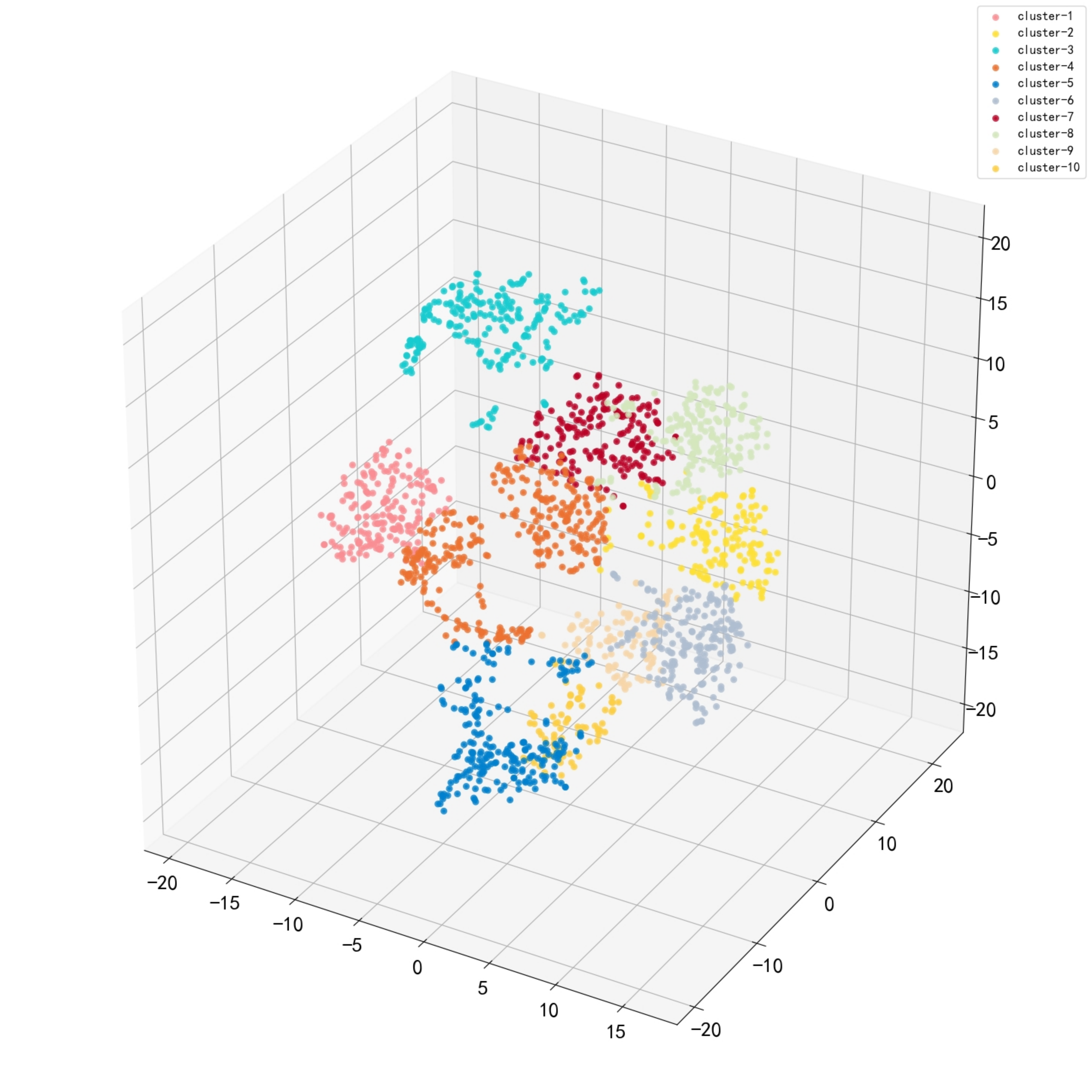}}
	\subfigure[RI-FCM
	]{
		\label{fig:subfig:Dataset3}
		\includegraphics[width=0.23\textwidth]{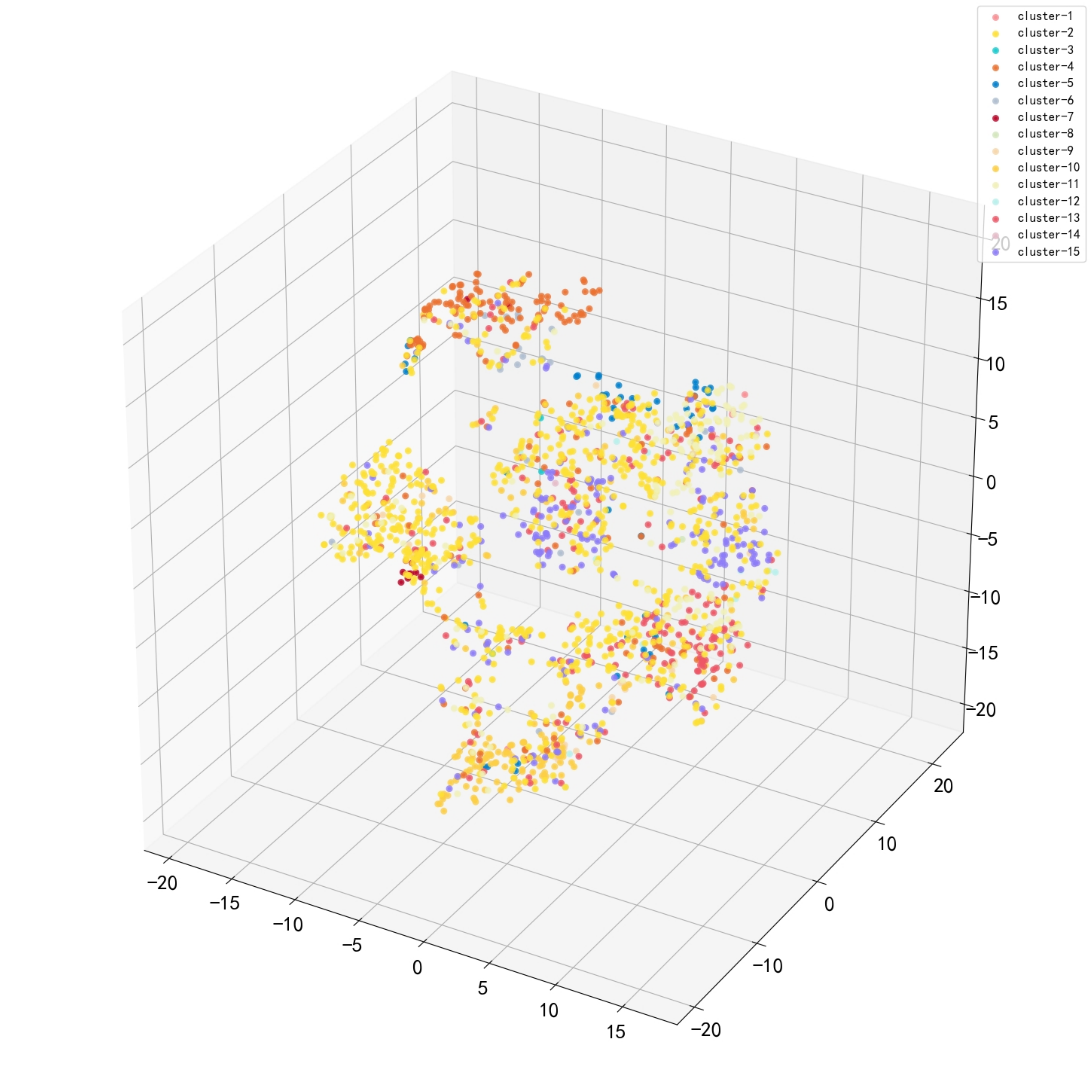}}
	\subfigure[FDPC
	]{
		\label{fig:subfig:Dataset3}
		\includegraphics[width=0.23\textwidth]{mnist-ecm-moead.pdf}}
	\caption{The clustering result of other different methods  on  mnist dataset.}
	\label{fig:syn}
\end{figure*}


\begin{table*}[htbp]
	\centering
	\caption{The comparison of ACC, NMI and ARI of algorithms on real data sets}
	 \setlength{\tabcolsep}{1mm}{
\begin{tabular}{cccccccccccc}
	\hline
	Dataset              & Index & FGB\_overlap   & FGB\_K-means   & K-means & FCM            & ECM-NSGA-II & ECM-MOEA/D & MFS-FCM & DI-FCM & RL-FCM & FDPC           \\ \hline
	& ARI   & \textbf{0.915} & 0.905          & 0.902   & 0.879          & 0.614       & 0.447      & 0.902   & 0.786  & 0.301  & 0.666          \\
	mnist                & NMI   & \textbf{0.934} & 0.924          & 0.921   & 0.906          & 0.805       & 0.727      & 0.921   & 0.864  & 0.568  & 0.8            \\
	& ACC   & 0.950          & \textbf{0.954} & 0.953   & 0.934          & 0.694       & 0.571      & 0.953   & 0.822  & 0.505  & 0.737          \\ \hline
	& ARI   & \textbf{0.773} & 0.715          & 0.718   & 0.604          & 0.297       & 0.317      & 0.653   & 0.625  & 0.651  & 0.644          \\
	cell                 & NMI   & \textbf{0.927} & 0.894          & 0.895   & 0.845          & 0.573       & 0.596      & 0.873   & 0.857  & 0.536  & 0.612          \\
	& ACC   & 0.831          & \textbf{0.837} & 0.836   & 0.699          & 0.398       & 0.390      & 0.751   & 0.728  & 0.511  & 0.661          \\ \hline
	& ARI   & 0.568          & \textbf{0.592} & 0.583   & 0.551          & 0.567       & 0.564      & 0.583   & 0.583  & 0.453  & 0.568          \\
	Iris                 & NMI   & \textbf{0.734} & 0.641          & 0.635   & 0.612          & 0.695       & 0.699      & 0.635   & 0.635  & 0.569  & 0.734          \\
	& ACC   & 0.667          & \textbf{0.813} & 0.807   & 0.764          & 0.726       & 0.722      & 0.807   & 0.807  & 0.767  & 0.667          \\ \hline
	& ARI   & 0.531          & \textbf{0.605} & 0.563   & 0.571          & 0.473       & 0.504      & 0.605   & 0.605  & 0.455  & 0.349          \\
	Soybean              & NMI   & 0.732          & \textbf{0.740} & 0.723   & 0.730          & 0.608       & 0.647      & 0.74    & 0.74   & 0.571  & 0.635          \\
	& ACC   & 0.617          & \textbf{0.809} & 0.766   & 0.726          & 0.686       & 0.638      & 0.809   & 0.809  & 0.574  & 0.532          \\ \hline
	\multicolumn{1}{l}{} & ARI   & 0.211          & \textbf{0.469} & 0.445   & 0.433          & 0.261       & 0.233      & 0.356   & 0.357  & 0.319  & 0.257          \\
	dermatology          & NMI   & 0.457          & \textbf{0.597} & 0.562   & 0.509          & 0.481       & 0.460      & 0.564   & 0.562  & 0.521  & 0.582          \\
	\multicolumn{1}{l}{} & ACC   & 0.508          & \textbf{0.514} & 0.511   & 0.503          & 0.465       & 0.476      & 0.439   & 0.444  & 0.419  & 0.411          \\ \hline
	\multicolumn{1}{l}{} & ARI   & \textbf{0.391} & 0.034          & 0.049   & 0.061          & 0.067       & 0.038      & 0.046   & 0.047  & 0.049  & 0.034          \\
	banknote             & NMI   & \textbf{0.469} & 0.020          & 0.030   & 0.040          & 0.067       & 0.093      & 0.029   & 0.029  & 0.291  & 0.098          \\
	\multicolumn{1}{l}{} & ACC   & \textbf{0.634} & 0.596          & 0.612   & 0.619          & 0.595       & 0.606      & 0.609   & 0.610  & 0.106  & 0.603          \\ \hline
	\multicolumn{1}{l}{} & ARI   & 0.206          & \textbf{0.223} & 0.185   & 0.180          & 0.155       & 0.203      & 0.180   & 0.169  & 0.114  & 0.222          \\
	ukm                  & NMI   & 0.292          & 0.296          & 0.269   & 0.263          & 0.190       & 0.249      & 0.264   & 0.263  & 0.193  & \textbf{0.300} \\
	\multicolumn{1}{l}{} & ACC   & 0.485          & \textbf{0.562} & 0.500   & 0.475          & 0.505       & 0.534      & 0.484   & 0.481  & 0.426  & 0.504          \\ \hline
	\multicolumn{1}{l}{} & ARI   & 0.094          & 0.191          & 0.186   & \textbf{0.197} & 0.113       & 0.123      & 0.186   & 0.189  & 0.065  & 0.134          \\
	segment              & NMI   & 0.210          & \textbf{0.349} & 0.336   & 0.340          & 0.299       & 0.295      & 0.336   & 0.339  & 0.329  & 0.242          \\
	\multicolumn{1}{l}{} & ACC   & 0.273          & \textbf{0.372} & 0.367   & 0.368          & 0.288       & 0.307      & 0.368   & 0.369  & 0.135  & 0.331          \\ \hline
\end{tabular}
}
	
	\label{tab:synr}%
\end{table*}%

\begin{table*}[htbp]
	\centering
	\caption{The comparison of running time of algorithms on real  datasets(unit:second)}
	 \setlength{\tabcolsep}{1mm}{
	\begin{tabular}{ccccccccccc}
		\hline
		Dataset     & FGB\_overlap   & FGB\_K-means    & K-means        & FCM            & ECM-NSGA-II & ECM-MOEA/D & MFS-FCM  & DI-FCM  & RL-FCM   & FDPC    \\ \hline
		mnist       & 3.624          & \textbf{2.374}  & \textbf{0.067} & 8.309          & 33.416      & 27.019     & 41.443   & 19.190  & 135.895  & 32.076  \\
		cell        & 57.039         & \textbf{19.654} & \textbf{0.290} & 112.430        & 474.911     & 507.599    & 1107.395 & 629.280 & 2697.948 & 830.548 \\
		Iris        & 0.106          & \textbf{0.119}  & \textbf{0.022} & \textbf{0.053} & 7.330       & 6.093      & 0.246    & 0.594   & 0.578    & 0.234   \\
		Soybean     & 0.028          & 0.032           & \textbf{0.016} & \textbf{0.014} & 8.693       & 10.721     & 0.203    & 0.126   & 0.094    & 0.047   \\
		dermatology & 0.375          & \textbf{0.328}  & \textbf{0.063} & 0.674          & 12.252      & 16.744     & 2.469    & 3.547   & 5.689    & 1.625   \\
		banknote    & 3.644          & 1.784           & \textbf{0.037} & \textbf{0.484} & 14.506      & 28.931     & 3.374    & 3.233   & 178.996  & 18.988  \\
		ukm         & \textbf{0.214} & 0.236           & \textbf{0.038} & 0.216          & 11.646      & 11.273     & 1.987    & 2.804   & 2.249    & 0.678   \\
		segment     & 6.599          & \textbf{4.825}  & \textbf{0.089} & 6.533          & 34.941      & 34.676     & 9.544    & 56.599  & 390.204  & 57.082  \\ \hline
\end{tabular}
}
	
	\label{tab:synt}%
\end{table*}%

\subsection{ Experiments on Real Datasets }

For experiments on real datasets, first of all, we conducted experiments on the 3D real mnist dataset. 
Fig. 4(a) was the GroundTruth result of the original mnist dataset. And then Fig. 4(b) is the result of  fuzzy granular-balls generated by our method. As shown in Fig. 4(c) the cluster result generated by the algorithm FGB\_overlap we proposed, Fig. 4(d) is the cluster result generated by the algorithm FGB\_K-means that we propose. In Fig. 5(a)-5(h), we can see different clustering result with 8 different methods, such as K-means, FCM, ECM-NSGA-II, ECM-MOEA/D, MFS-FCM, DI-FCM, RI-FCM, FDPC on mnist dataset. It can be seen that clustering effects of algorithm FGB\_overlap and algorithm FGB\_K-means based  fuzzy granular-balls we proposed  are obviously better than other comparison algorithms, and the results are more fit the GroundTruth. At the same time, the results of some cluster assessment indicators can also explain this. According to Table II, we can  see that indicator ARI and NMI of our method FGB\_overlap  on dataset mnist are better than the other 8 algorithms, and the ACC indicator on the FGB\_K-means on dataset mnist is also the highest. Finally, from Table 4, we can also see that in terms of the running time of the program, the fastest K-means algorithm is removed, and our method time is also ranked in the top three on dataset mnist.

The second selected real dataset is the
biomedical image segmentation of immune cells in confocal
microscopy on a spatiotemporal biomedical dataset [39], [40].
This dataset is particularly relevant for images of immune
cells, which exhibit high plasticity. From Table II, we can  see that indicator ARI and NMI of our method FGB\_overlap  on dataset mnist are better than the other 8 algorithms, and the ACC indicator on the FGB\_K-means on dataset mnist is also the highest. Finally, from Table 4, we can also see that in terms of the running time of the program, the fastest K-means algorithm is removed, and our method time is also ranked in the top three on dataset mnist.

Further more, several real datasets from the UCI Machine Learning Repository [41] were used. The relevant information of these datasets is shown in Table I. 
The comparison of ACC, NMI and ARI scores are illustrated in Table II and the running time of these algorithms is shown in Table III. It can be seen from Table II that compared with the other 8 comparison algorithms, the evaluation index values (ARI, NMI, ACC) of our proposed FGB\_overlap and FGB\_K-means on the 6 UCI real datasets are mostly ranked the top two, demonstrate the effectiveness and accuracy of our method. In addition, based on Table III, we can also see that the running time of the FGB\_overlap and FGB\_K-means proposed by us on the 6 UCI real datasets can also rank in the top three, which also proves the efficiency and feasibility of our method.

\begin{figure}
	\centering
	\includegraphics[width=0.9\linewidth]{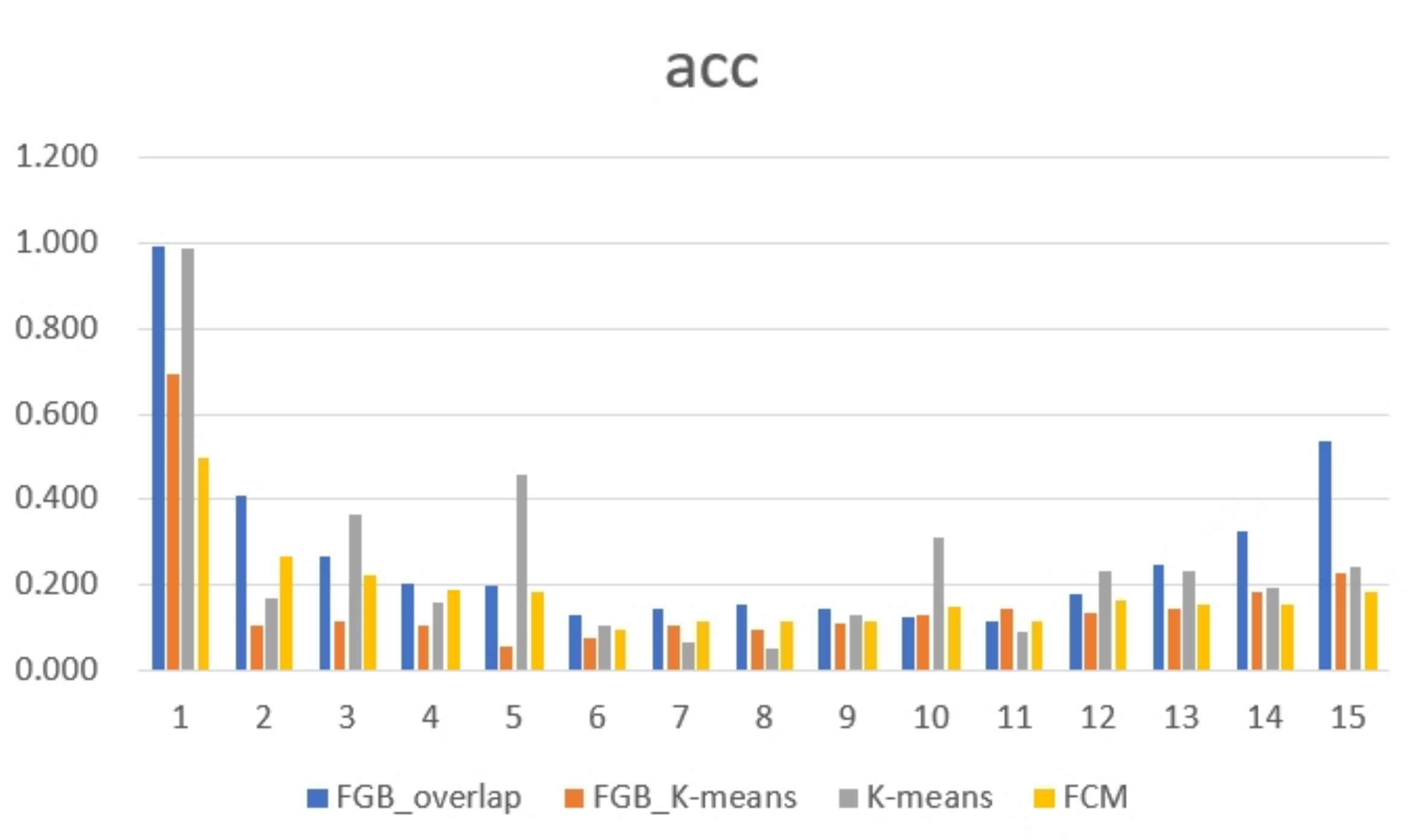}
	\caption[acc]{The comparison of acc of algorithms on simulated brain mr datasets}
	\label{fig:mr-acc}
\end{figure}

\subsection{ Experiments on Simulated Brain MR Datasets}
Finally, we also selected real datasets from the Simulated Brain Database (SBD) on the BrainWeb website for experiments,
Currently, the SBD contains simulated brain MRI data based on two anatomical models: normal and multiple sclerosis (MS).  For both of these, full 3-dimensional data volumes have been simulated using three sequences (T1-, T2-, and proton-density- (PD-) weighted) and a variety of slice thicknesses, noise levels, and levels of intensity non-uniformity.  Among these two models, we chose the second model, multiple sclerosis (MS), and we obtained their GroundTruth through appropriate pre-processing. The specific operation is to first set each option to T1 mode. Under the icmb protocol, the slice thickness is 1mm, the noise level is 0, and the uneven gray level is 0. Then select the first 150 layers of ms pathological brain images, and select only 0, 1, 2, 3, and 10 types of data for each layer for experiments, After segmentation, various types are rendered with the following pixel values. The last 10 layers are grouped into a group. Finally, 15 groups of GroundTruth and Label are obtained for experiment and verification

\begin{table}[htbp]
	\centering
	\caption{The comparison of running time of algorithms on simulated brain mr datasets(unit:second)}
	\setlength{\tabcolsep}{2mm}{
\begin{tabular}{ccccc}
	\hline
	group & FGB\_overlap & FGB\_K-means & K-means & FCM   \\ \hline
	1     & 0.101        & 0.118        & 0.014   & 0.152 \\
	2     & 1.110        & 1.958        & 0.021   & 1.793 \\
	3     & 2.962        & 5.013        & 0.027   & 3.070 \\
	4     & 3.837        & 7.080        & 0.031   & 4.017 \\
	5     & 5.138        & 12.168       & 0.037   & 5.190 \\
	6     & 4.709        & 14.302       & 0.046   & 7.273 \\
	7     & 4.846        & 13.883       & 0.042   & 7.118 \\
	8     & 5.624        & 14.594       & 0.043   & 7.021 \\
	9     & 4.533        & 12.781       & 0.043   & 6.661 \\
	10    & 3.703        & 11.048       & 0.038   & 5.823 \\
	11    & 3.481        & 9.668        & 0.038   & 5.136 \\
	12    & 4.260        & 8.292        & 0.034   & 4.757 \\
	13    & 4.315        & 6.697        & 0.030   & 4.031 \\
	14    & 6.045        & 7.474        & 0.024   & 6.244 \\
	15    & 0.938        & 1.161        & 0.017   & 1.010 \\ \hline
\end{tabular}}
	
	\label{tab:synt}%
\end{table}%

We first directly use k-means algorithm and original FCM algorithm to conduct experiments on these 15 sets of data sets, and count the experimental results. Then we use algorithm 1 to generate fuzzy hyperspheres of these 15 sets of data sets. On these fuzzy hyperspheres, we use k-means algorithm and original FCM algorithm to cluster, and count the experimental results. Finally, we use our own overlay method to connect the generated fuzzy hyperspheres to generate clustering results, As shown in Figure 3, the experimental results are statistically analyzed. It can be seen that the method of generating fuzzy hyperspheres by overlapping connection proposed by us has a relatively stable and good clustering effect. At the same time, the clustering accuracy of most datasets is also excellent. Most of its accs are also higher than k-means and FCM clustering methods. In addition, as shown in Figure 4, compared with the running time, we can see that the running time of k-means is the shortest, and the overlay method we proposed is also in a high efficiency position, which is still superior to the original FCM clustering method.

\section{CONCLUSION}

In this paper, we represent the appropriate granularity of the dataset by generating fuzzy granular-balls while characterizing the fuzzy properties of the data. The obtained fuzzy granular-balls can be used in various ways to complete the final clustering for different datasets scenarios. For example, using connection by overlap can adaptively obtain clustering results of complex shapes; using K-means can obtain clustering results more quickly.  Experiments on synthetic datasets and real datasets show that FGB\_overlap and FGB\_K-means can recognize patterns in complex datasets, and it is also very effective. There are still many exploration directions for future research. In the future, we will explore how to improve the original FCM iteration method and connect more possibilities of fuzzy granular-balls to further improve the effect and efficiency of the algorithm.

\section{ACKNOWLEDGMENTS}
This work was supported in part by the National Natural
Science Foundation of China under Grant Nos. 62176033 and 61936001, the Natural Science Foundation of Chongqing
under Grant Nos. cstc2019jcyj-msxmX0485 and cstc2019jcyj-
cxttX0002 and by NICE: NRT for Integrated Computational
Entomology, US NSF award 1631776

\ifCLASSOPTIONcaptionsoff
  \newpage
\fi

\end{document}